\documentclass{article}

\usepackage{graphicx}
\usepackage{amssymb}
\usepackage{epstopdf}
\usepackage[tight,footnotesize]{subfigure}
\usepackage{stfloats}
\usepackage{microtype}
\usepackage{units}
\usepackage{algorithmic}
\usepackage{algorithm}
\usepackage{url}

\usepackage{float}
\begin{document}

\title{Simulated Data Experiments for Time Series Classification Part 1: Accuracy Comparison with Default Settings}
\author{Anthony Bagnall, Aaron Bostrom, James Large and Jason Lines\\
School of Computing Sciences \\
              University of East Anglia    \\
              Norwich NR4 7TJ              \\
              United Kingdom\\
              ajb@uea.ac.uk
              }

\maketitle

\begin{abstract}
There are now a broad range of time series classification (TSC) algorithms designed to exploit different representations of the data. These have been evaluated on a range of problems hosted at the UCR-UEA TSC Archive (www.timeseriesclassification.com), and there have been extensive comparative studies. However, our understanding of why one algorithm outperforms another is still anecdotal at best. This series of experiments is meant to help provide insights into what sort of discriminatory features in the data lead one set of algorithms that exploit a particular representation to be better than other algorithms. We categorise five different feature spaces exploited by TSC algorithms then design data simulators to generate randomised data from each representation. We describe what results we expected from each class of algorithm and data representation, then observe whether these prior beliefs are supported by the experimental evidence. We provide an open source implementation of all the simulators to allow for the controlled testing of hypotheses relating to classifier performance on different data representations. We identify many surprising results that confounded our expectations, and use these results to highlight how an over simplified view of classifier structure can often lead to erroneous prior beliefs. We believe ensembling can often overcome prior bias, and our results support the belief by showing that the ensemble approach adopted by the Hierarchical Collective of Transform based Ensembles (HIVE-COTE) is significantly better than the alternatives when the data representation is unknown, and is significantly better than, or not significantly significantly better than, or not significantly worse than, the best other approach on three out of five of the individual simulators.
\end{abstract}

\section{Introduction}

This technical report describes code to generate simulated time series classification (TSC) problems and presents an experimental evaluation of ten TSC algorithms on a range of simulation settings. Detailed algorithmic background can be found in~\cite{bagnall16bakeoff}. This research is part of our goal to try to determine why some TSC algorithms work better than others on different types of problems with the ultimate aim of having better {\em a priori} estimates of what algorithm will be the most accurate for a new classification problem.

The simulation code is part of an extensive TSC codebase\footnote{UEA TSC Repository {\tt https://bitbucket.org/TonyBagnall/time-series-classification}}, and is in the package \texttt{statistics.simulators}. Examples on how to use the simulators are in class \texttt{examples.SimulationExperiments}. There are two base structures of the simulations code: the \texttt{Model} class and the \texttt{DataSimulator} class. \texttt{DataSimulator} contains a \texttt{Model} for each problem class and is used to generate data. There are three basic use cases to generate a simulated data set.

\begin{enumerate}
\item  Set up models externally then call \texttt{generateData}.

\begin{verbatim}
ArrayList<Model> m = ....
DataSimulator ds = new DataSimulator(m);
Instances data=ds.generateData();
\end{verbatim}

\item Use a subclass of \texttt{DataSimulator}
\begin{verbatim}
DataSimulator ds = new SimulateShapeletDataset();
Instances data=ds.generateData();
\end{verbatim}
\item Use a bespoke static method in a subclass of \texttt{DataSimulator}
\begin{verbatim}
 data=SimulateShapeletData.generateShapeletData(seriesLength,casesPerClass);
\end{verbatim}
\end{enumerate}

Each subclass of \texttt{DataSimulator} has its own set of parameters, which can be passed as a 2-D array. Otherwise defaults will be used. the \texttt{DataSimulator} has parameters with the following defaults:
\begin{verbatim}
    int nosClasses=2;
    int seriesLength=100;
    int nosPerClass=50;
    int[] casesPerClass=new int[]{nosPerClass,nosPerClass};
\end{verbatim}
these can all be set by modifier methods. Default settings are in method \texttt{examples.SimulationExperiments.setStandardGlobalParameters}.

Currently implemented model subclasses are \texttt{ShapeletModel, DictionaryModel, ArmaModel, IntervalModel, ElasticModel} and \texttt{WhiteNoiseModel}.

\begin{figure}[!ht]
	\centering
       \includegraphics[height=10cm,  trim={2cm 5cm 2cm 5cm},clip]{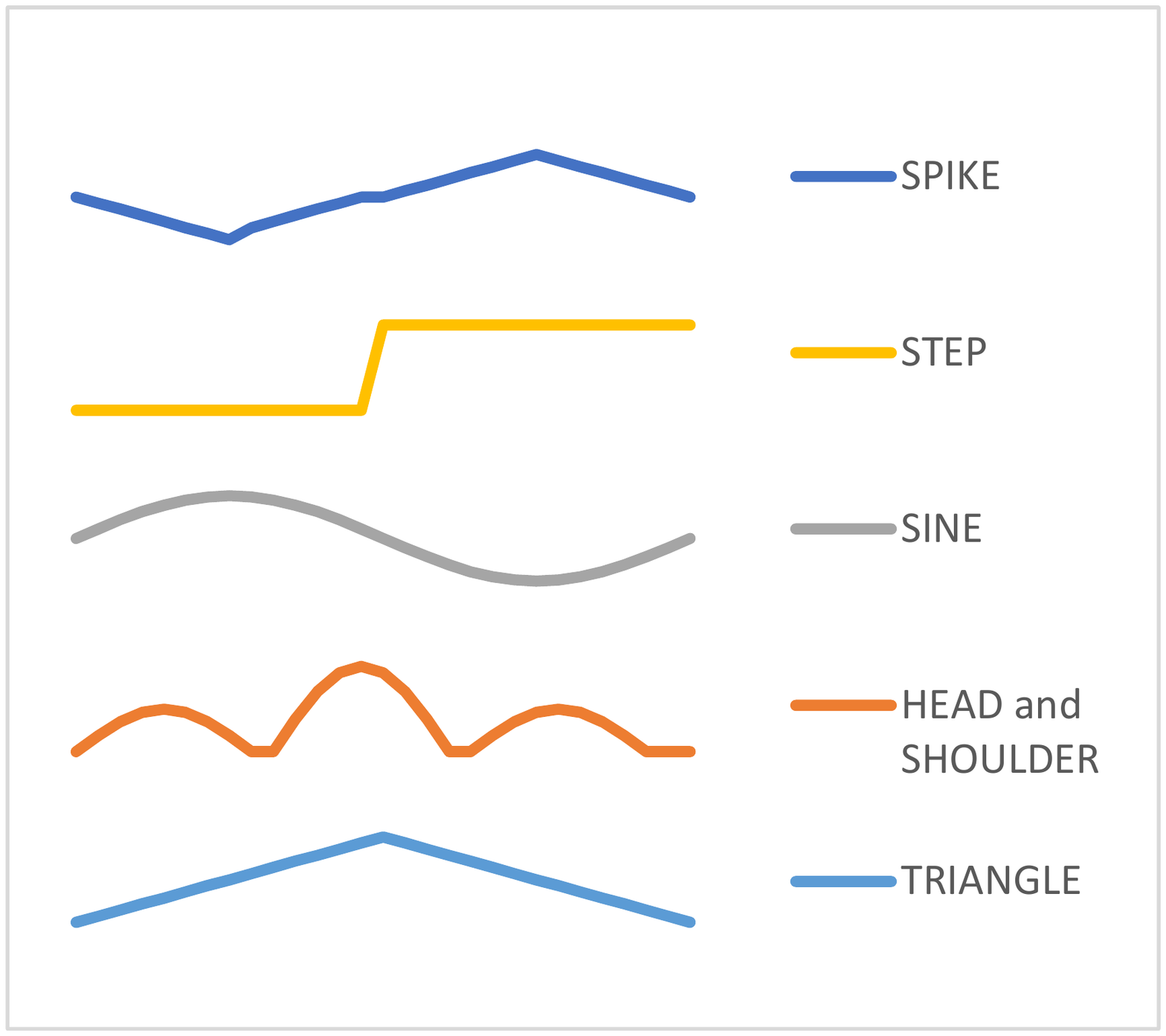}              	\caption{The five types of shapes that are used in all of the simulator models. }
       \label{shapes}
\end{figure}

Each simulator is based on placing one or more of the shapes shown in Figure~\ref{shapes} on a (usually longer) white noise series. The location, size, frequency and/or type of shape define the simulator.

\section{Algorithms Summary}

There have been many approaches for solving TSC problems proposed in the literature. Due to the rich and diverse nature of these solutions, we believe the best way to understand them is to group algorithms by the nature of the discriminatory features that they use. A more extensive description can be found in~\cite{bagnall16bakeoff}.

\subsection{Traditional Classification Approach}

It is possible, and indeed often sensible, to treat TSC problems as standard classification problems, i.e. to not explicitly use the fact the attributes are ordered. As a general principle, we advocate always starting with a traditional classifier. This acts as a sanity check and provides evidence that the more complex TSC approaches are in fact of value for a specific problem. We evaluate using two vector based classifiers: Rotation Forest (RotF)~\cite{rodriguez06rotation} and a heterogeneous ensemble of standard classifiers (HESCA)~\cite{lines16hive}.

\subsection{Whole series elastic}

Whole series techniques compare two series by a distance measure that uses all data points, then classify with a Nearest Neighbour (NN) approach. The simplest such approach is to compare series using Euclidean Distance. However, this baseline is easily beaten in practice, and most research effort has been directed at finding techniques that can compensate for small misalignments between series using {\bf elastic} distance measures. The almost universal benchmark for whole series measures is Dynamic Time Warping (DTW). Numerous alternatives have been proposed. These involve alternative warping criteria~\cite{jeong11weighted}, using versions of edit distance~\cite{marteau09stiffness,stefan13msm} and transforming to use first order differences~\cite{gorecki14nonisometric,batista14_cid}. We use two elastic distance classifiers. 1-NN DTW with window size set through cross validation (DTW)~\cite{ratanamahatana05threemyths} and the Elastic Ensemble (EE)~\cite{lines15elastic},  an ensemble of thirteen different 1-NN elastic classifiers.

\subsection{Intervals}

Rather than use the whole series, the interval class of algorithm select one or more phase-dependent intervals of the series. At its simplest, this involves a feature selection of a contiguous subset of attributes. However, the three most effective techniques generate multiple intervals, each of which is processed and forms the basis of a member of an ensemble classifier~\cite{deng13forest,baygogan13tsbf,baydogan2015time}. We use the Time Series Forest (TSF)~\cite{deng13forest} to represent interval based classifiers, because in~\cite{bagnall16bakeoff} it was shown to be not significantly different in accuracy to the alternatives and it is considerably simpler and faster.

\subsection{Shapelets}

Shapelet~\cite{Ye2011shapelets} approaches are a family of algorithms that focus on finding patterns that define a class and can appear anywhere in the series. A class is distinguished by the presence or absence of one or more shapelets somewhere in the whole series. Two common ways of finding shapelets are through enumerating the candidate shapelets in the training set~\cite{hills14shapelet} or searching the space of all possible shapelets with a form of gradient descent~\cite{grabocka14lst}. We use the balanced Shapelet Transform in conjunction with HESCA (ST-HESCA)~\cite{hills14shapelet,bostrom15binary} because in~\cite{bagnall16bakeoff} it was shown to be significantly more accurate than the alternatives.

\subsection{Dictionary-based}

Some problems are distinguished by the frequency of repetition of subseries, rather than by their presence or absence. Dictionary-based methods form frequency counts of recurring patterns, then build classifiers based on the resulting histograms~\cite{lin12bagofpatterns,schafer15boss}. We use Bag of SFA Symbols (BOSS)~\cite{schafer15boss} to represent dictionary classifiers, because in~\cite{bagnall16bakeoff} it was shown to be significantly more accurate than the alternatives.

\subsection{Spectral}

The frequency domain will often contain discriminatory information that is hard to detect in the time domain. Methods include constructing an autoregressive model~\cite{corduas2008time,bagnall14histogram} or combinations of autocorrelation, partial autocorrelation and autoregressive features~\cite{bagnall15cote}. The Random Interval Spectral Ensemble~\cite{lines16hive} is a way of combing spectral features in a decision tree ensemble and is used to represent Spectral classifiers.

\subsection{Combinations of the previous}
These distinctions are not absolute, and some classifiers could be assigned to multiple groups. For example, BOSS uses a Fourier transform, so could be classed as Spectral, and RISE uses random intervals, so could come in the Interval category. We differentiate between this inevitable fuzziness in algorithm classification and algorithms that explicitly use features that combine two or more of the above approaches to representation into a single classifier. Two examples of this combination approach are presented in~\cite{kate16dtw_features}, where a classifier is built on concatenated different feature spaces, and~\cite{fulcher14comparative} which uses forward selection of features for a linear classifier. However, the most accurate method for combining approaches is the Collective of Transform based Ensembles (COTE)~\cite{bagnall15cote} which involves transformation into a feature space that represents each group and ensembling classifiers together. The original version of COTE, Flat-COTE~\cite{bagnall15cote}, involves ensembling 35 classifiers in the elastic, shapelet and spectral domains into a single ensemble. More recently, Hierarchical COTE (HIVE-COTE)~\cite{lines16hive} adopts a more structured modular approach by encapsulating classifiers on each representation in a single ensemble then hierarchically ensembling.

\subsection{Evaluation}

An experiment consists of a number of repetitions of the process of randomly generating a data set, splitting it into train and test sets, building the classifier on the train (including all parameter optimisation) then evaluating once on the train set. We tabulate the results for the mean accuracy, the standard error in accuracy and the mean rank for the ten algorithms listed in Table~\ref{classifiers}..
To test the accuracy of multiple classifiers over multiple randomly generated independent datasets, we start with a Friedman's test to detect for a significant difference between any classifiers, then identify where significant differences occur by constructing a critical difference diagram as described by \cite{demsar06comparisons}. However, following recommendations in (\cite{benavoli16posthoc}) and (\cite{garcia08pairwise}), we have abandoned the Nemenyi post-hoc test used in (\cite{demsar06comparisons}) to form groups of classifiers within which there is no significant difference (cliques). Instead, we compare all classifiers with pairwise Wilcoxon signed rank tests, and form cliques using the Holm correction, which adjusts family-wise error less conservatively than a Bonferonni adjustment. It is worthwhile pointing out that cliques formed this way do not necessarily reflect the rank order. For example, if we have three classifiers $(A,B,C)$ with average ranks $(A>B>C)$, it is possible for A to be significantly worse than $B$ but not significantly worse than $C$ in pairwise tests. This relationship cannot easily be displayed on a critical difference diagram. Happily, we did not encounter this phenomena with any of the results we present in this paper.

\begin{table}
\caption{TSC algorithms used in experimentation}
\label{classifiers}
\resizebox{\columnwidth}{!}{
\begin{tabular}{lll} \hline
Algorithm                                               & Acronym & Type\\ \hline
Rotation Forest~\cite{rodriguez06rotation}              & RotF	  & Vector\\
Heterogenous Ensemble of Standard Classifiers~\cite{lines16hive}            & HESCA   & Vector  \\
Dynamic Time Warping                                    & DTW	  & Elastic\\
Elastic Ensemble~\cite{lines15elastic}                  & EE	  & Elastic\\
Time Series Forest~\cite{deng13forest}                  & TSF     & Interval\\
Shapelet Transform~\cite{hills14shapelet}               & ST-HESCA	   & Shapelet\\
Bag of SFA Symbols~\cite{schafer15boss}                 & BOSS    & Dictionary (Spectral)      \\
Random Interval Spectral Ensemble~\cite{lines16hive}                         & RISE	  & Spectral (Intervals)\\
Flat Collective of Transform based Ensembles~\cite{bagnall15cote} & Flat-COTE & Combination\\
Hierarchical COTE~\cite{lines16hive}  & HIVE-COTE & Combination\\
\hline
\end{tabular}
}
\end{table}

\section{Whole Series Elastic Simulator: \\ \texttt{ElasticModel} and \texttt{SimulateElasticData}}

\begin{figure}[!ht]
	\centering
\begin{tabular}{cc}
       \includegraphics[height =6cm, trim={2cm 4cm 2cm 6cm},clip]{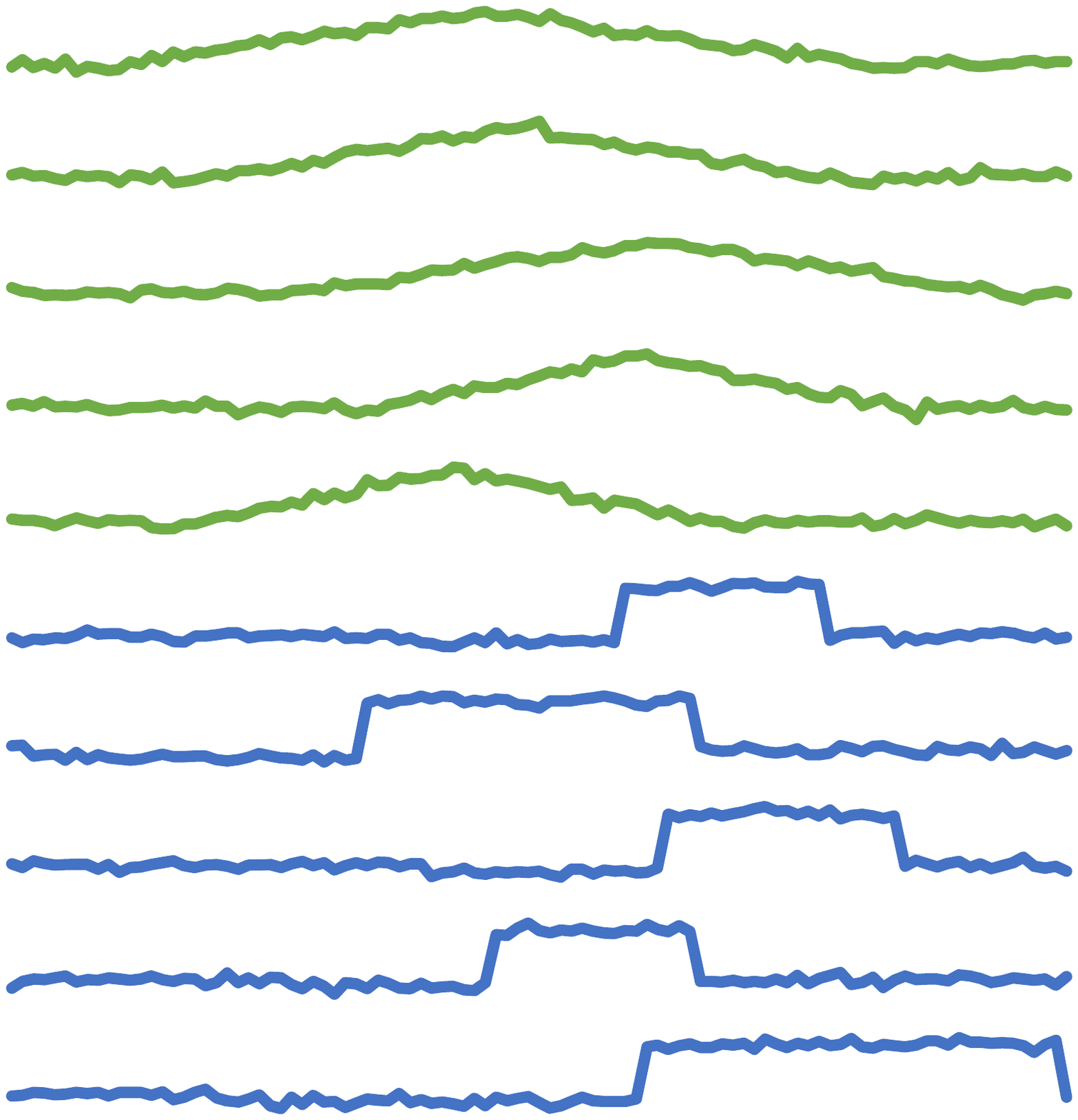}              	
&
       \includegraphics[height=6cm,trim={2cm 4cm 2cm 6cm},clip]{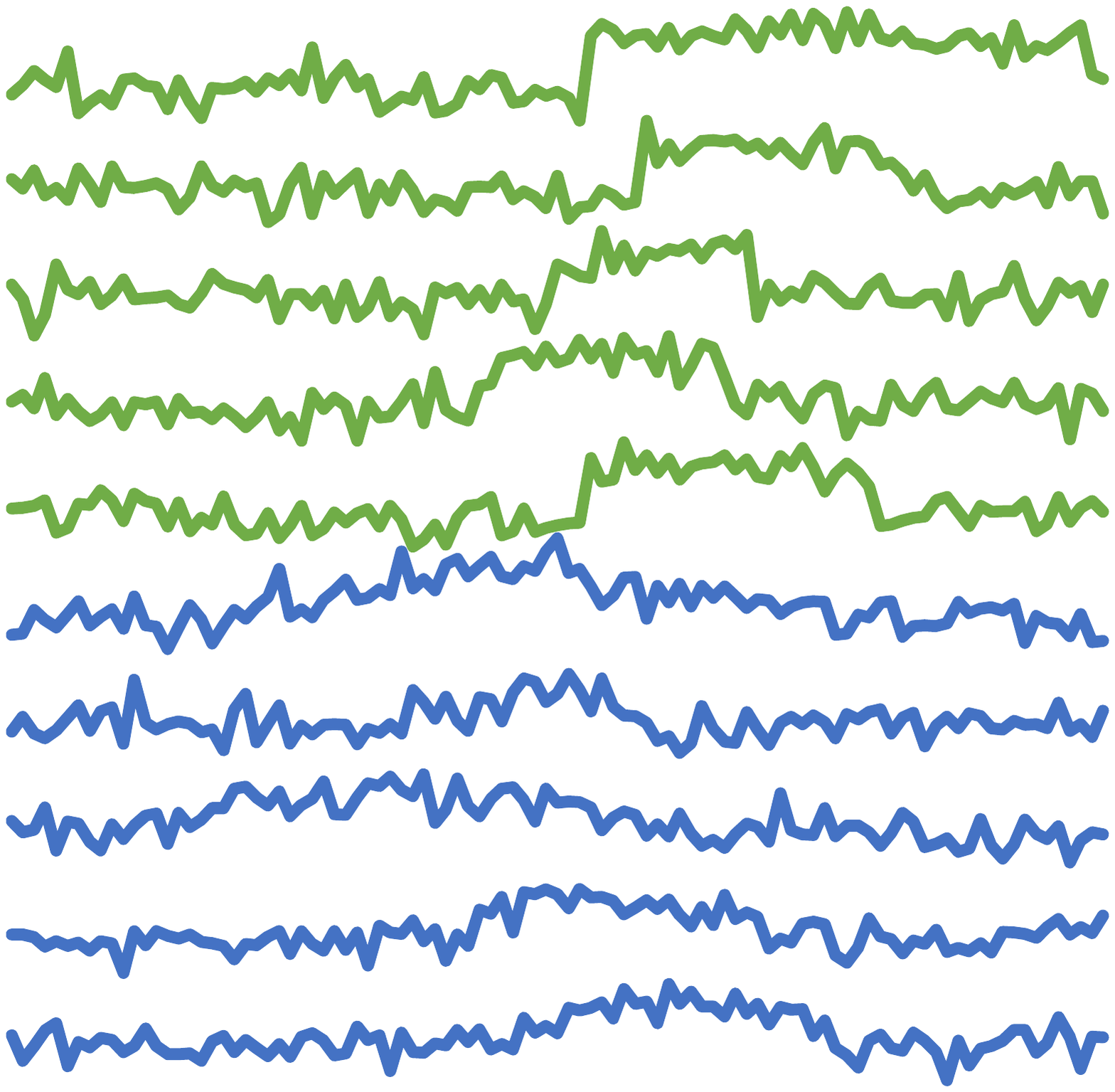}              	\\
       (a) & (b) \\
\end{tabular}
       \caption{Two examples of simulated elastic data with a triangle shape (top 5) and a step shapelet (bottom 5) stretched to different degrees. Figure (a) has low noise and Figure (b) has standard white noise. }
       \label{elasticEx}
\end{figure}

Elastic techniques are based on an elastic distance measure between two series. The measures are called elastic because they attempt to compensate for misalignments between otherwise similar series. The elastic simulator models this scenario by using a single shape (from those shown in Figure~\ref{shapes}) to define each class. Variation within the class is achieved by altering the length of the shape for each series. Elastic classifiers should be able to compensate for this stretching, but it may confound other techniques. For each instance of the class the length of the shape is randomly selected to be between 20\% and 100\% of the total series length. Example series with low and standard noise are shown in Figure~\ref{elasticEx}.

\subsection{Prior Belief}
We have the following prior beliefs concerning performance on this data:
\begin{enumerate}
\item We think vector based classifiers (RotF and HESCA) would be the worst approach for this type of data;
\item Shapelet classifiers (ST-HESCA) should be confounded by the variable length of shapes, but may do better than vector based classifiers;
\item  Interval classifiers (TSF) may do well because of the averaging/smoothing they perform over intervals. This may also help dictionary classifiers (BOSS);
\item We would expect the elastic measures to do best, with EE coming out on top; and
\item We would not expect combined approaches (Flat-COTE and HIVE-COTE) to produce much benefit.
\end{enumerate}

\subsection{Results}

\begin{figure}[!ht]
	\centering
       \includegraphics[width=12cm,trim={4cm 12cm 4cm 11.5cm},clip=true]{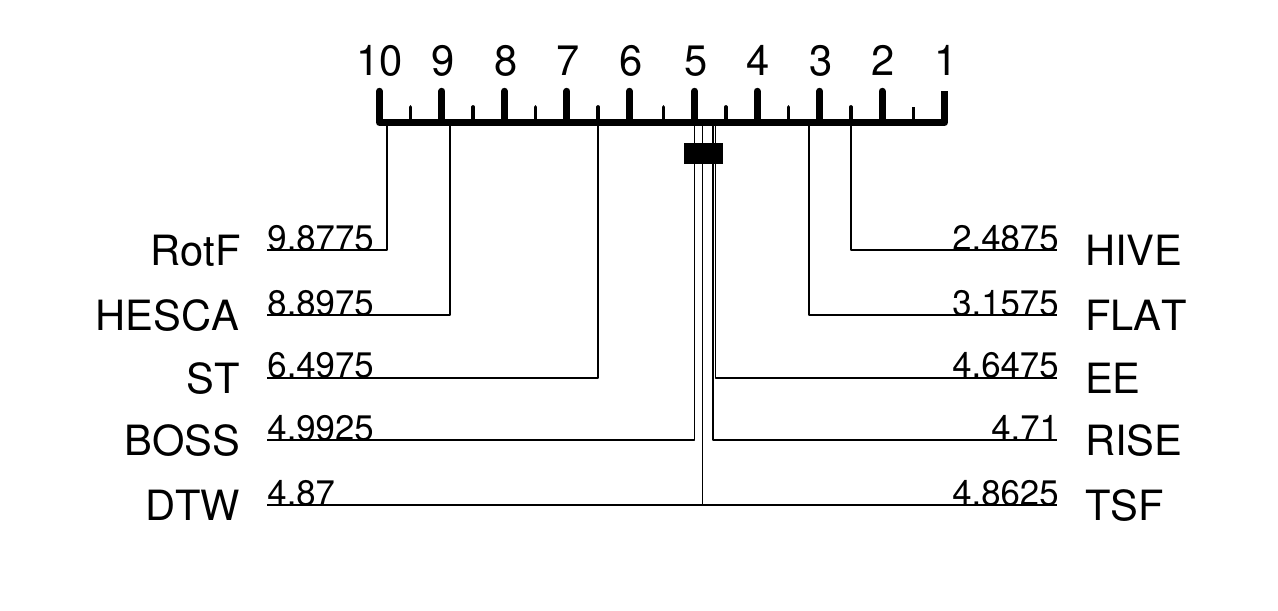}              	
       \caption{Critical difference diagram for ten classifiers over 200 elastic simulations. }
       \label{elasticCD}
\end{figure}

\begin{figure}[!ht]
	\centering
       \includegraphics[width=12cm,trim={4cm 10cm 4cm 10cm},clip]{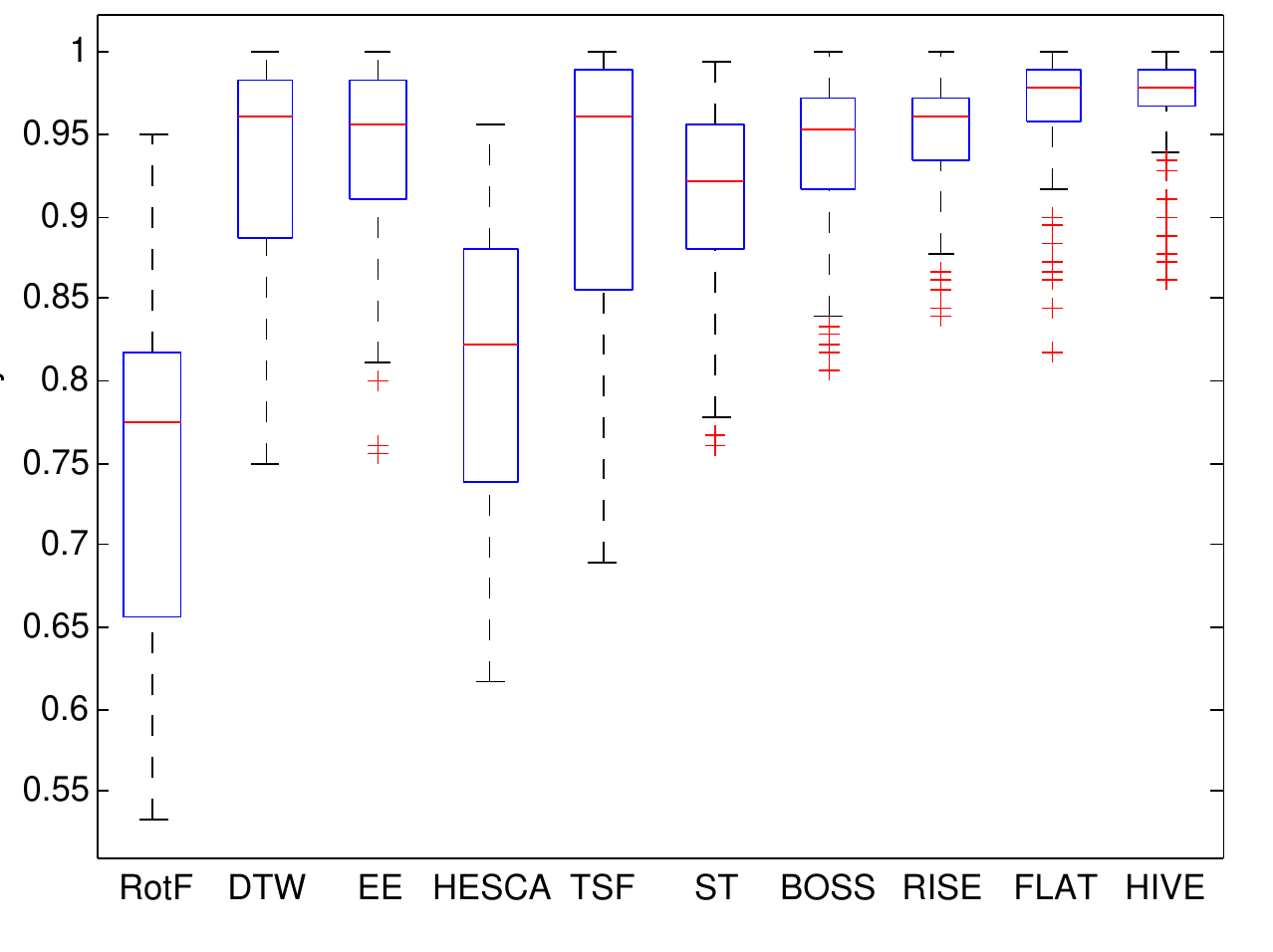}              	
       \caption{Box plots for ten classifiers over 200 elastic simulations. }
       \label{elasticBoxPlot}
\end{figure}

The method \texttt{setStandardGlobalParameters(String str)} sets the global parameters for all simulators based on the argument \texttt{str}. These parameters have been set to balance difficulty against time/space complexity. For the elastic simulator, the parameters are as follows:

\begin{verbatim}
        seriesLength=100;
        nosCases={100,100};
        propTrain=0.1;
\end{verbatim}

Thus the train set size is 20, the test size 180. We generate 200 random models with these parameters and evaluate the ten classifiers listed in Table~\ref{classifiers}. The results are presented in Table~\ref{elasticAcc}. The critical difference diagram is shown in Figure~\ref{elasticCD}. To demonstrate the range of complexities of the problem, the box plots of the 200 resamples are shown in Figure~\ref{elasticBoxPlot}.

\begin{table}
\centering
\caption{Mean Accuracy of ten classifiers averaged over 200 different random elastic datasets.}
\label{elasticAcc}
\begin{tabular}{l|c|c|c} \hline
Algorithm & Mean Accuracy & Standard Error & Average Rank \\ \hline
HIVECOTE  &  97.29\% &	0.19\% & 2.49\\
FlatCOTE  &  96.85\% &	0.22\% & 3.16\\
RISE      &  95.17\% &	0.23\% &  4.71\\
EE	      &  93.99\% & 0.36\%  &  4.65\\
BOSS	  &  93.94\% & 0.32\%  & 4.99      \\
DTW	      &  93.45\% &	0.42\% &   4.87      \\
TSF	      &  92.36\% & 0.53\%  &  4.86       \\
ST-HESCA	      &  91.35\% & 0.38\%  &  6.50\\
HESCA	  &  80.83\% & 0.58\%  &  8.90\\
RotF	  &  74.46\% & 0.67\%  & 9.88\\ \hline
\end{tabular}
\end{table}

\subsection{Post Experiment Observations}

We make the following observations about the results presented in Table~\ref{elasticAcc} and Figures~\ref{elasticCD} and~\ref{elasticBoxPlot}.\\

\noindent{\bf Vector based}. As expected, the vector based approaches (RotF and HESCA) are clearly the worst approaches for this kind of data. Both algorithms are significantly worse than all of the others. RotF is significantly worse than HESCA, which only  demonstrates that ensembling bad classifiers can make them better, but still worse than a more sensible approach. This confirms our a priori belief.\\

\noindent{\bf Shapelet}. ST-HESCA stands alone as better than the vector based but significantly worse than all the others. It finds some information in shape, but not enough to make it as good as the other approaches.\\

\noindent{\bf Elastic, interval and dictionary}. Surprisingly, there is no difference at all between DTW, EE, TSF, RISE and BOSS. This contradicts our prior beliefs, but with hindsight we can make a case  for why this happens. Consider TSF and RISE; both take random intervals, then take summary measures over the interval. If we compare a triangle to a step, for example, they have regions where they are more likely to have a larger value or slope. It would require a much more complex, in depth analysis to work out what is actually going on with these algorithms, perhaps by deconstructing how they actually classify. Whilst that would be interesting, our current goal is to perform a black box analysis of results. There are several interesting features in Table~\ref{elasticAcc}. For example, RISE has an average accuracy of 95\%, which is higher than the 94\% EE achieves. however, EE has a higher average rank, which suggests the elastic approaches fail badly on some configurations. This is demonstrated by the fact that EE has a wider range in the box plot (Figure~\ref{elasticBoxPlot}) than RISE. Broadly, we believe that the equivalence of performance is caused by their being two confounding factors in the data: warping of the shape and random noise. DTW and EE can compensate for warping, but are confounded by noise. BOSS, RISE and TSF all involve some form of averaging and smoothing, and are thus better at coping with random noise. Our conclusion is that is dangerous to put too much credence on prior beliefs as to the best approach for a problem.\\

\noindent{\bf Combined}. Both HIVE and Flat are significantly better than all the other algorithms. They are taking advantage of the unexpected performance of TSF, RISE and BOSS to find a better classifier. These results demonstrate the benefit of not trusting our prior beliefs too much, and allowing an ensemble classifier to automatically balance different approaches. HIVE is significantly better than Flat, which is unsurprising, since unlike Flat it includes both TSF and BOSS. The box plot shows that HIVE has a much tighter band of accuracies resulting in much less variance.  HIVE-COTE can to some degree compensate against our prior bias as to the superiority of one algorithm over another.



\section{Interval Simulator \\ \texttt{IntervalModel} and \texttt{generateIntervalData}}

\begin{figure}[!ht]
	\centering
\begin{tabular}{cc}
       \includegraphics[height =6cm, trim={2cm 2cm 2cm 2cm},clip]{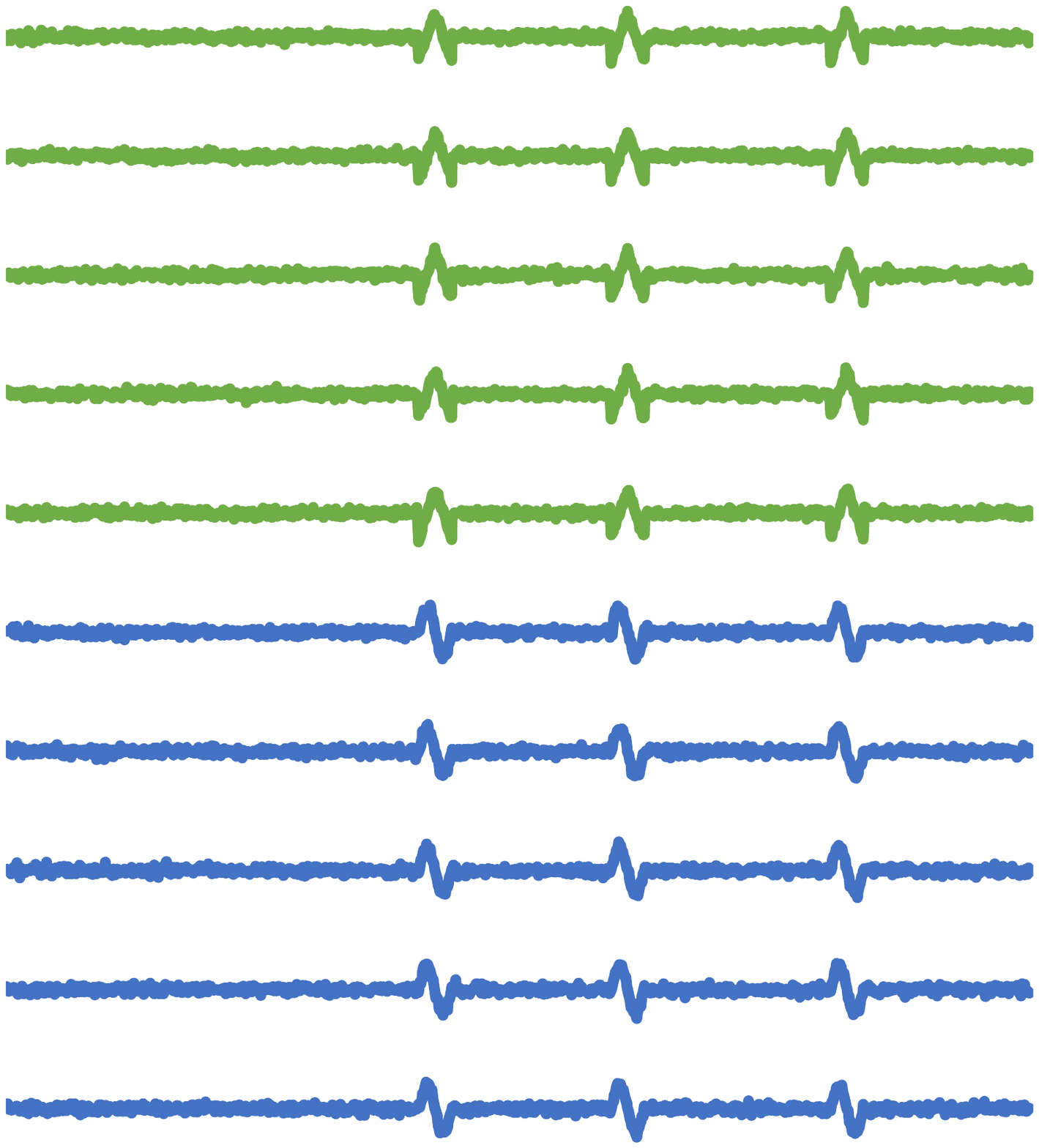}              	
&
       \includegraphics[height=6cm,trim={2cm 2cm 2cm 2cm},clip]{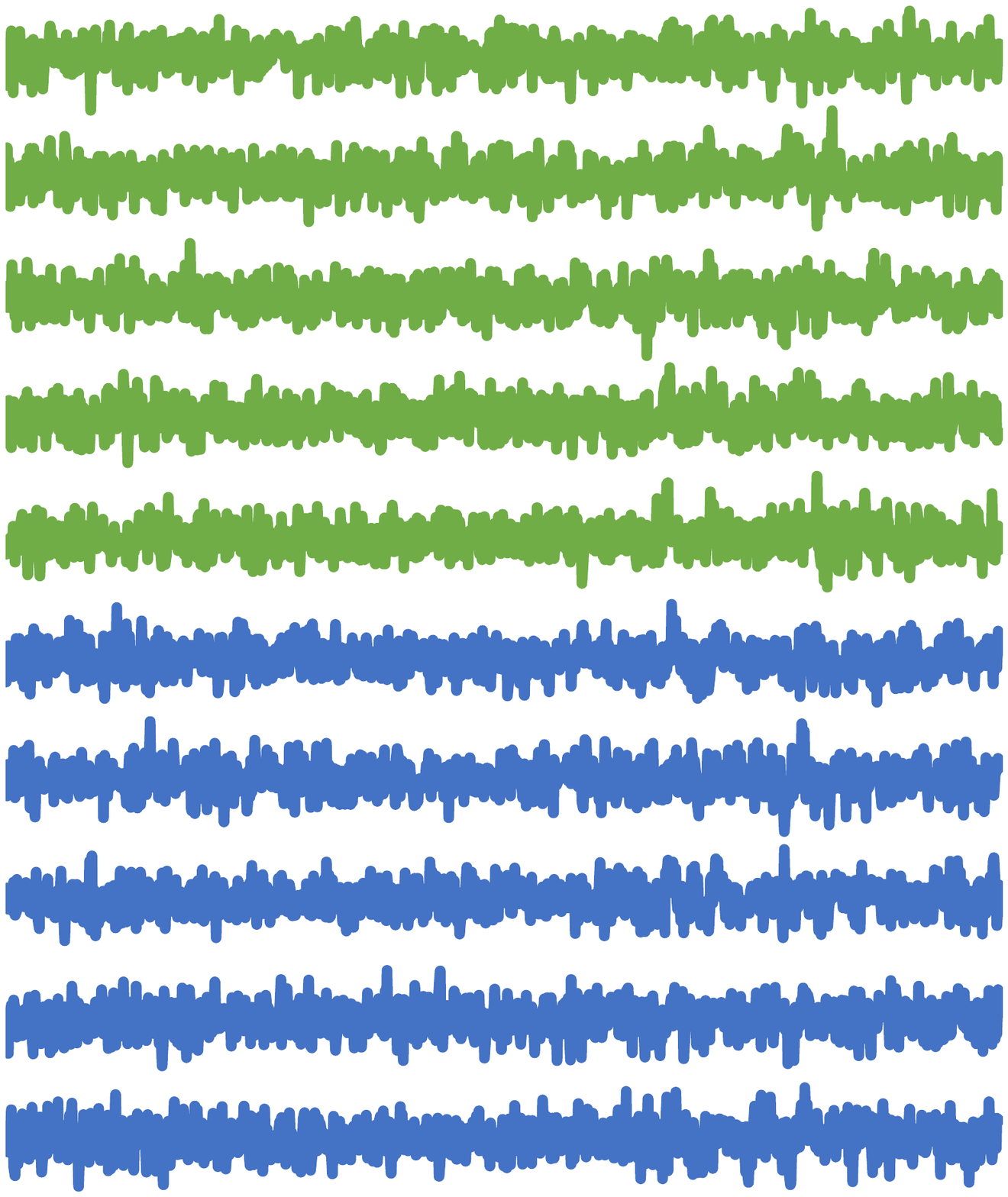}              	\\
       (a) & (b) \\
\end{tabular}
       \caption{Examples of simulated data for interval classifiers with head and shoulders (top 5) and spike shapes (bottom 5).  Figure (a) has low noise and Figure (b) has standard white noise. }
       \label{intervalEx}
\end{figure}

Interval classifiers take multiple intervals across the series, calculate some features from the interval, then ensemble classifiers over the intervals. One scenario where this may be effective is when there are localised discriminatory patterns which appear at the same location in every series, but are much rarer than the random noise in which they are embedded. To model this scenario we insert a fixed number of short shapes in long series of noise. The class is defined by a single shape, repetitions of which appear at fixed locations in all series. The confounding factor here is purely the amount of noise. Figure~\ref{intervalEx} gives some example series.

\subsection{Prior Belief}
We have the following prior beliefs concerning performance on the interval data:
\begin{enumerate}
\item we think vector based classifiers (RotF and HESCA) would be quite good at these data, because we have in affect created a standard classification problem with a large amount of noise;
\item we would expect the elastic classifiers (DTW and EE) to be confounded by the level of noise;
\item Shapelet classifiers (ST-HESCA) and dictionary classifiers (BOSS) may do well, because we have not mixed up the shapelets between intervals within a class. The simulator should probably be redesigned to help confound ST-HESCA and BOSS;
\item  Interval classifiers (TSF) should be best at these problems; and
\item Combination classifiers will probably not be better than interval classifiers, but should be at least as good.
\end{enumerate}

\subsection{Results}
\begin{figure}[!ht]
	\centering
       \includegraphics[width=12cm,trim={4cm 12cm 4cm 11.5cm},clip]{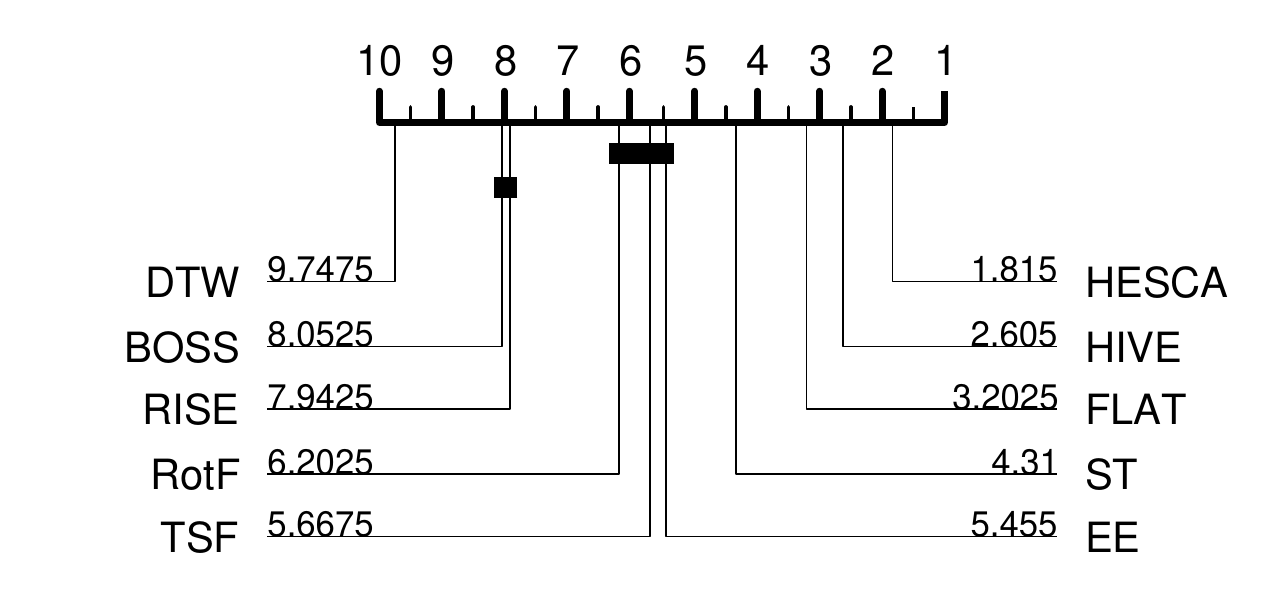}              	
       \caption{Critical difference diagram for ten classifiers over 200 interval simulations. }
       \label{intervalCD}
\end{figure}

\begin{figure}[!ht]
	\centering
       \includegraphics[width=12cm,trim={3.5cm 10cm 3.5cm 10cm},clip]{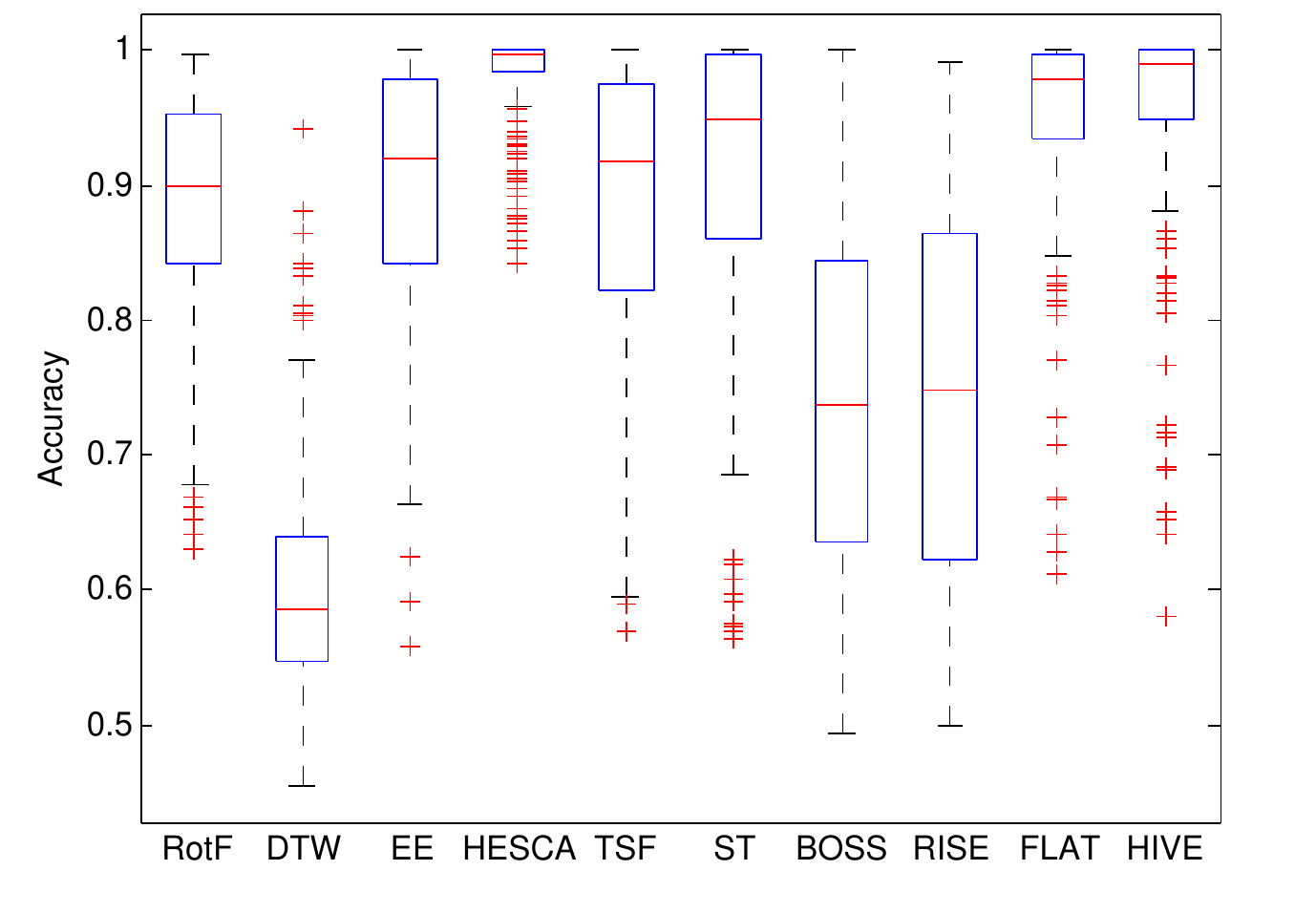}              	\caption{Box plots for ten classifiers over 200 interval simulations. }
       \label{intervalBoxPlot}
\end{figure}

The default parameters are:
\begin{verbatim}
    seriesLength =1000;
    int[] nosCases={200,200};
    double trainProp=0.1;
    int numIntervals=3;
    int shapeToNoiseRatio=10;//Determines the length of the intervals
\end{verbatim}
The series are longer for this simulation, and 901 of the 100 features are white noise. Each interval is length 33.
The results are presented in Table~\ref{intervalAcc}. The critical difference diagram is shown in Figure~\ref{intervalCD}. To demonstrate the range of complexities of the problem, the box plots of the 200 resamples are shown in Figure~\ref{intervalBoxPlot}.

\begin{table}
\centering
\caption{Mean Accuracy of ten classifiers averaged over 200 different random interval datsets.}
\label{intervalAcc}
\begin{tabular}{l|c|c|c}\hline
Algorithm & Mean Accuracy & Standard Error & Mean Rank \\ \hline
HESCA       & 97.95\% & 0.29\%  & 1.75\\
HIVECOTE    & 95.42\% & 0.70\%  & 3.05\\
FlatCOTE    & 94.86\% & 0.66\%  & 3.90\\
ST-HESCA          & 91.03\% & 0.92\%  & 3.59\\
EE          & 89.64\% & 0.72\%  & 5.41\\
TSF         & 88.85\% & 0.77\%  & 6.12\\
RotF        & 88.12\% & 0.60\%  & 5.59\\
BOSS        & 74.30\% & 1.00\%  & 7.54\\
RISE        & 74.09\% & 0.92\%  & 8.18\\
DTW         & 60.49\% & 0.38\%  & 9.83\\ \hline
\end{tabular}
\end{table}

\subsection{Post Experiment Observations}

The interval simulator results are also surprising. Our expectation that TSF and RISE would be the best was not born out by the results. We make the following observations about the results presented in Table~\ref{intervalAcc} and Figures~\ref{intervalCD} and~\ref{intervalBoxPlot}.\\

\noindent{\bf Vector based}. HESCA is significantly better than all other classifiers, including the other vector based classifier RotF (which is a constituent of HESCA). Both facts surprised us. HESCA is designed as a component for other classifiers which reduces the variance of accuracy estimate without decreasing the actual accuracy. On extensive experiments reported in~\cite{lines16hive} it was found to be not significantly less accurate than RotF on 72 UCI data sets, whilst producing less biased accuracy estimates from cross validation on the train data. However, we did not expect it to be significantly better than RotF on this data. This suggests two things. Firstly, HESCA may in general be a good approach for problems with a high proportion of redundant features. Secondly, although RotF is usually good at handling redundant and noisy features, in the scenario when the number of noise features is high and the discriminatory features are contiguous, it struggles. This is a result of the fact that the selection of random features is unlikely to capture elements of the same shape in the same tree. So, for example, the peak of a triangle and a head and shoulders are only really distinguishable when other points in the same shape are also taken into account. We believe this merits further investigation. \\

\noindent{\bf Spectral and dictionary}. RISE and BOSS are significantly better than DTW, not significantly different to each other, and significantly worse than all the others. Both algorithms are hampered by the relatively small size of the interval. RISE will not be able to capture the autocorrelation within shapes because they are so short. In principle, BOSS should capture the class differences because each shape occurs in only one class. However, the spectral smoothing used by BOSS suffers from the same problems as RISE.\\

\noindent{\bf Elastic and interval}. DTW is the worst algorithm on this data, but there is no significant difference between RotF, TSF and EE. The poor performance of DTW is not unexpected, since there is no benefit from realigning the series and a large chance of confounding the classifier with noise. The fact that EE is significantly better is surprising. We believe this is due to the edit distance classifiers contained within EE, which we think will be better at compensating for sections of random noise. The other alternative is that the difference based classifiers within EE better capture the discriminatory features. This could be tested through further experimentation, but it is not a high priority because EE is outperformed by other classifiers.

The fact there is no difference between RotF, TSF and EE is also surprising. Post hoc, we believe that this is caused by the fact we used relatively short intervals for the shapes, and the averaging over random intervals introduces too much noise. It is interesting to consider the difference in approach between RotF and TSF. For each tree, RotF takes a random set of attributes, transform into principle components, then constructs a tree. TSF samples multiple intervals for each tree, taking the average, standard deviation of each interval. Neither approach successfully captures the discriminatory features for this type of simulation. \\

\noindent{\bf Shapelet}. ST-HESCA is significantly better than all approaches except HESCA and COTE. The discriminatory features are shapelets and pos hoc we realise the relatively good performance of ST-HESCA is an artifact of the fact that each class only uses a single shape over all intervals. It is worth noting that ST-HESCA is significantly better than BOSS, which should also benefit from the single shape per class. The fact that ST-HESCA uses shapes from the data makes it slow, but this is an example of the benefits from doing so.\\

\noindent{\bf Combined}. Both versions of COTE are significantly better than all except HESCA, and HIVE-COTE is significantly better than Flat-COTE. Neither COTE contains time domain classifiers such as those used by HESCA. This demonstrates how the COTE approach can achieve significant improvement through combining diverse but fairly weak, classifiers.

\section{Shapelet Simulator \\ \texttt{ShapeletModel} and \texttt{generateShapeletData}}

\begin{figure}[!ht]
	\centering
\begin{tabular}{cc}
       \includegraphics[height =6cm, trim={2cm 2cm 2cm 2cm},clip]{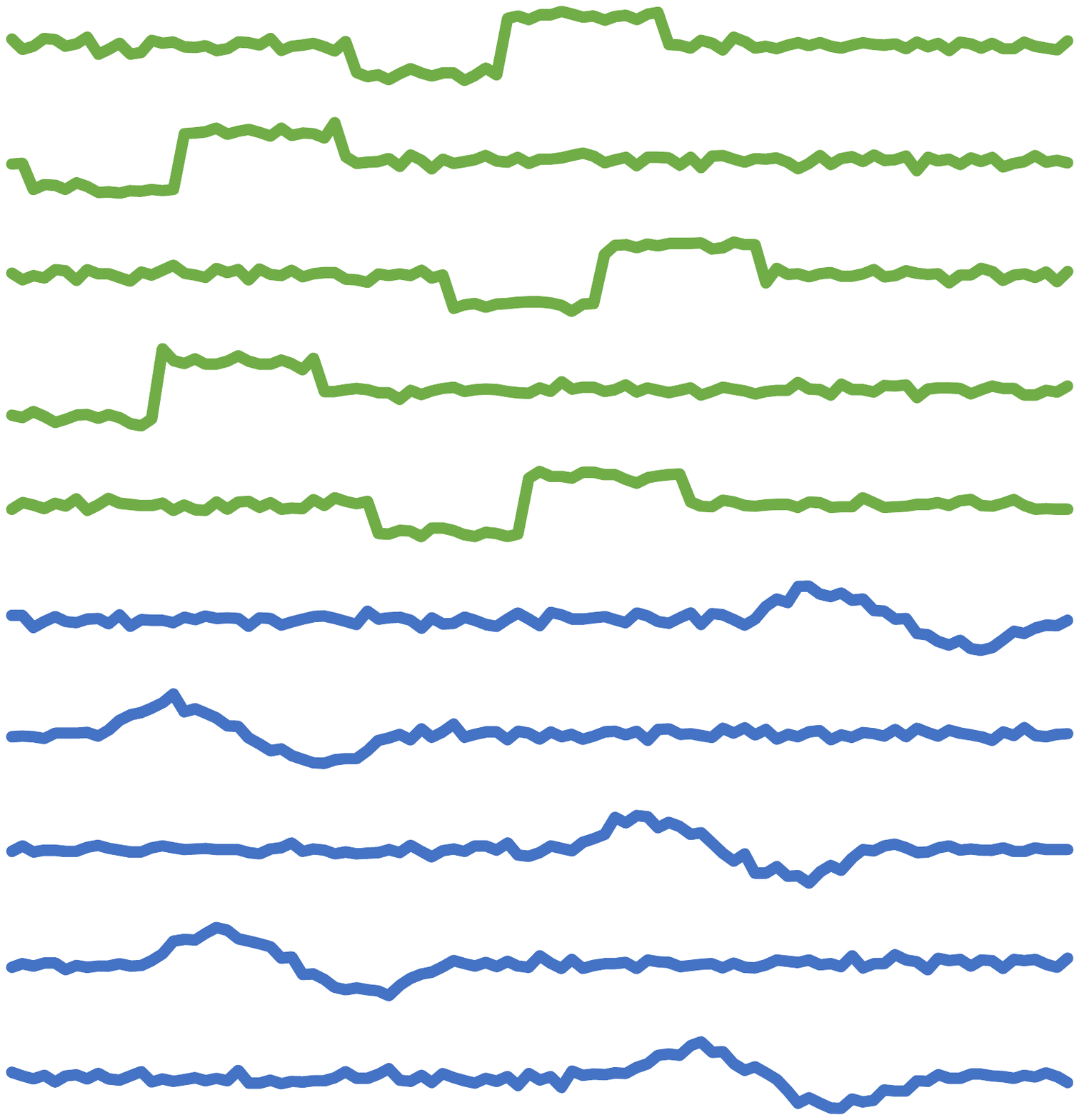}              	
&
       \includegraphics[height=6cm,trim={2cm 2cm 2cm 2cm},clip]{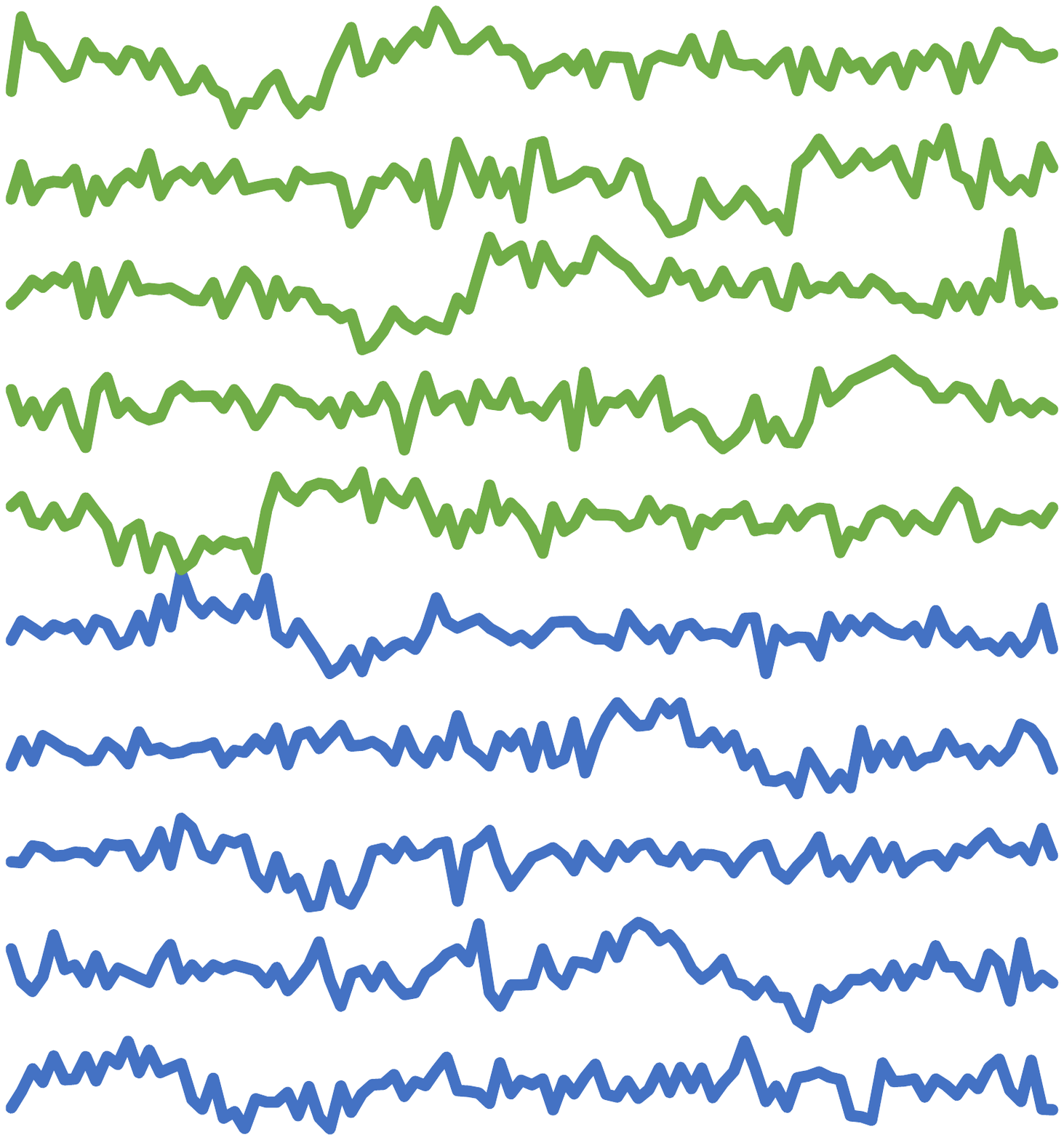}              	\\
       (a) & (b) \\
\end{tabular}
       \caption{Two examples of simulated data with a step shapelet (top 5) and a sine shapelet (bottom 5). Figure (a) has low noise and Figure (b) has standard white noise. }
       \label{shapeletEx}
\end{figure}

A shapelet is a discriminatory pattern embedded in a series. Currently this simulator only works for a single shapelet defining a class. Each model randomly chooses a shapelet from those described in Figure~\ref{fig1}. The shapelet is placed at a random location for each series from that model (by the call to \texttt{generateSeries(int length)}), embedded in noise. Figure~\ref{shapeletEx} gives some example series.

\subsection{Prior Belief}
We have the following prior beliefs concerning performance on the shapelet data:
\begin{enumerate}
\item we think vector based classifiers would be poor at these data, because the random location of the discriminatory shape should confound them;
\item interval classifiers should be poor at this data for the same reasons as the vector based classifiers;
\item the fact that a single shapelet defines the class means that we would expect the elastic classifiers to be fairly effective, because the warping may be able to compensate for the random location;
\item dictionary classifiers may also do well, but the presence or absence of a single shape will not always be apparent because of the discretisation stage.
\item shapelet classifiers should be the best, and combination classifiers will probably not be better than shapelet classifiers, but should be at least as good.
\end{enumerate}

\subsection{Results}
\begin{figure}[!ht]
	\centering
       \includegraphics[width=12cm,trim={4cm 12cm 4cm 11.5cm},clip]{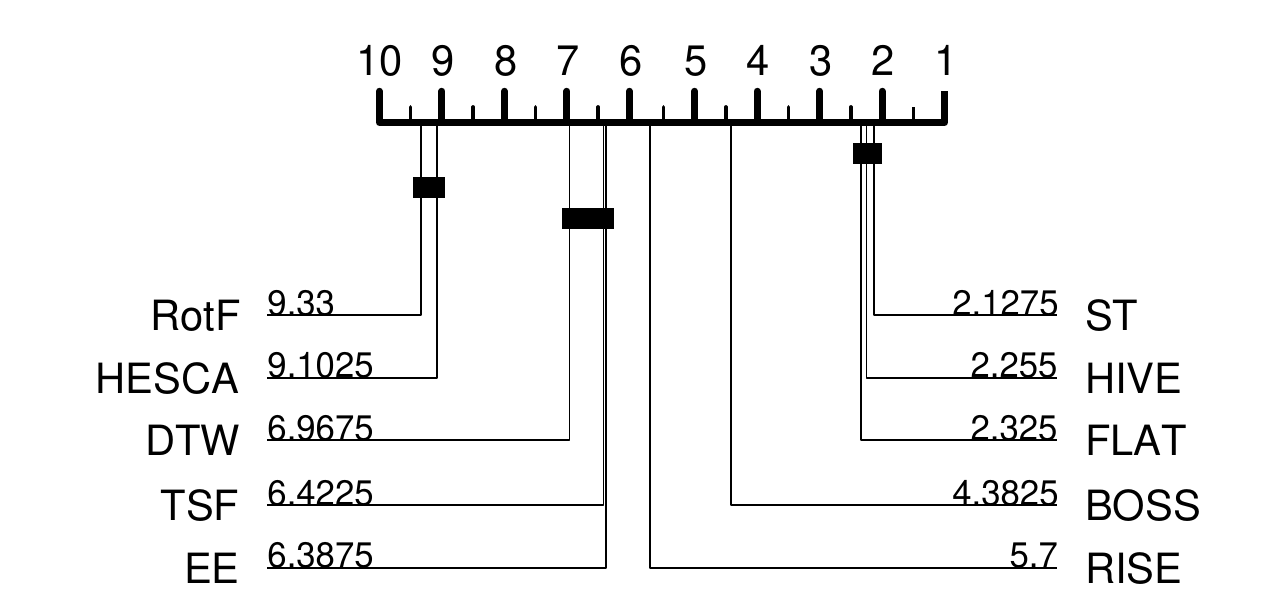}              	
       \caption{Critical difference diagram for ten classifiers over 200 shapelet simulations. }
       \label{shapeletCD}
\end{figure}

\begin{figure}[!ht]
	\centering
       \includegraphics[width=12cm,trim={3.5cm 10cm 3.5cm 10cm},clip]{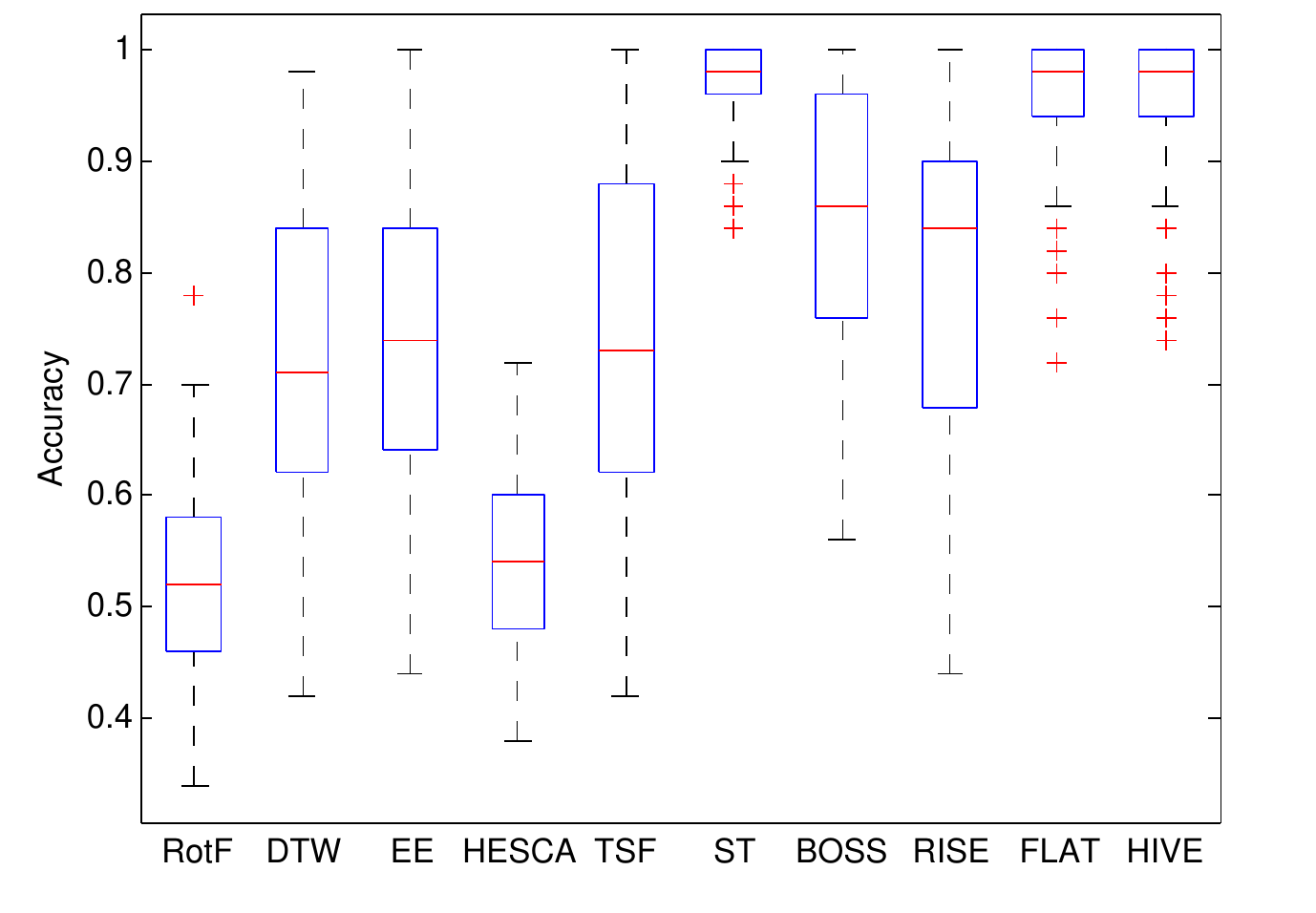}              	\caption{Boxplot for ten classifiers over 200 shapelet simulations. }
       \label{shapeletBoxPlot}
\end{figure}

The parameters (and default values) are:

\begin{verbatim}
        int numShapelets=1;
        int shapeletLength =29;
        int seriesLength=300;
        int[] nosCases={50,50};
\end{verbatim}

The series are shorter than for the interval classifier, but there is a similar shape to noise ratio: approximately ten times more noise features than shape features. The results are presented in Table~\ref{shapeletAcc} and Figures~\ref{shapeletCD} and ~\ref{shapeletBoxPlot}.These results can be recreated with the method \texttt{SimulationExperiment.runShapeletSimulatorExperiment()}.

\begin{table}
\centering
\caption{Accuracy of ten classifiers averaged over 100 different random shapelet problems.}
\label{shapeletAcc}
\begin{tabular}{l|c|c|c} \hline
Algo    &	Acc  &	Std Error  & Mean Rank\\ \hline
ST-HESCA          & 96.86\%    &  0.27\%    & 2.13  \\
HIVE-COTE   & 96.31\%    &  0.31\%    & 2.26  \\
Flat-COTE   & 95.74\%    &  0.37\%    & 2.33  \\
BOSS        & 85.35\%    &  0.83\%    & 4.38  \\
RISE        & 79.03\%    &  0.89\%    & 5.70  \\
TSF         & 74.04\%    &  1.05\%    & 6.42  \\
EE          & 73.96\%    &  0.94\%    & 6.39  \\
DTW         & 71.71\%    &  0.96\%    & 6.97  \\
HESCA       & 53.58\%    &  0.53\%    & 9.10  \\
RotF        & 51.26\%    &  0.54\%    & 9.33  \\  \hline
\end{tabular}
\end{table}

\subsection{Post Experiment Observations}
Broadly, the results are in line with our expectations. \\

\noindent{\bf Vector based}.
Both RotF and HESCA are little better than random guessing.\\

\noindent{\bf Elastic and interval}. There is no significant difference between DTW, TSF and EE. They are significantly better than the vector classifiers,
but worse than all other approaches. There is no significant difference between TSF and EE on the Elastic, Shapelet and Interval simulations. This raises the question of whether they are in fact discovering the same features, and that one is just a proxy for the other. \\

\noindent{\bf Spectral}. For reasons that are not immediately apparent, RISE does significantly better than the vector, elastic and interval methods and relatively better on shapelet data than on interval data. On reflection, we can make a case for this. RISE chooses a random interval for each ensemble member, then uses the spectral features to construct a classifier. Within each interval, the features are phase independent. Hence, if RISE picks an interval containing a shape, it may characterise that shape independent of where it is within the interval. If the interval is short, it will capture the shape well, but that is unlikely to be the location of the shape for new instances. If the interval is short, the noise will confound the shape characterisation, but it is more likely the shape of a new case will be within the interval. This trade off explains why it does not do as well as dictionary or shapelet approaches. \\

\noindent{\bf Dictionary}. BOSS is significantly better than RISE, but significantly worse than ST-HESCA and COTE. In principle, BOSS should be good at these problems. The windowed histograms should reflect that the word associated with a shape is in series of one class but not another. However, the frequency difference between histograms is just one, so it is possible that noise will overwhelm this difference.\\

\noindent{\bf Shapelet}.
ST-HESCA is significantly better than all apart from the COTE classifiers, of which it is a component. It was slightly surprising that it was only about 97\% accurate. Further experiments with different levels of noise have we think revealed why. ST-HESCA does not only find the true shapelet. It often finds shapes with only part of the true shape, because these can be as useful for classification as the full shape. This can cause a problem with a particular scenario. Suppose all the train set shapes are at least $x$ time steps from the beginning. Shapes with $y<x$ trailing points and the true shapelet will dichotomise the train data as well as the true shape, and thus will be included in the transform. It may even be the case that the true shape is not retained because there are so many of these padded shapes. Now if a training instance happens to have the true shape at the very beginning of the series, all of these shapelets with trailing points will be poor classification features, and this may confound the classifier. We are investigating mitigating against this unusual expceptional case.\\

\noindent{\bf Combined}. Both Flat-COTE and HIVE-COTE contain ST-HESCA, and both are able to perform as well as ST-HESCA despite the inclusion of poor classifiers.

\section{Dictionary Simulator \\ \texttt{DictionaryModel} and \texttt{generateDictionaryData}}

\begin{figure}[!ht]
	\centering
\begin{tabular}{cc}
       \includegraphics[height =6cm, trim={2cm 2cm 2cm 2cm},clip]{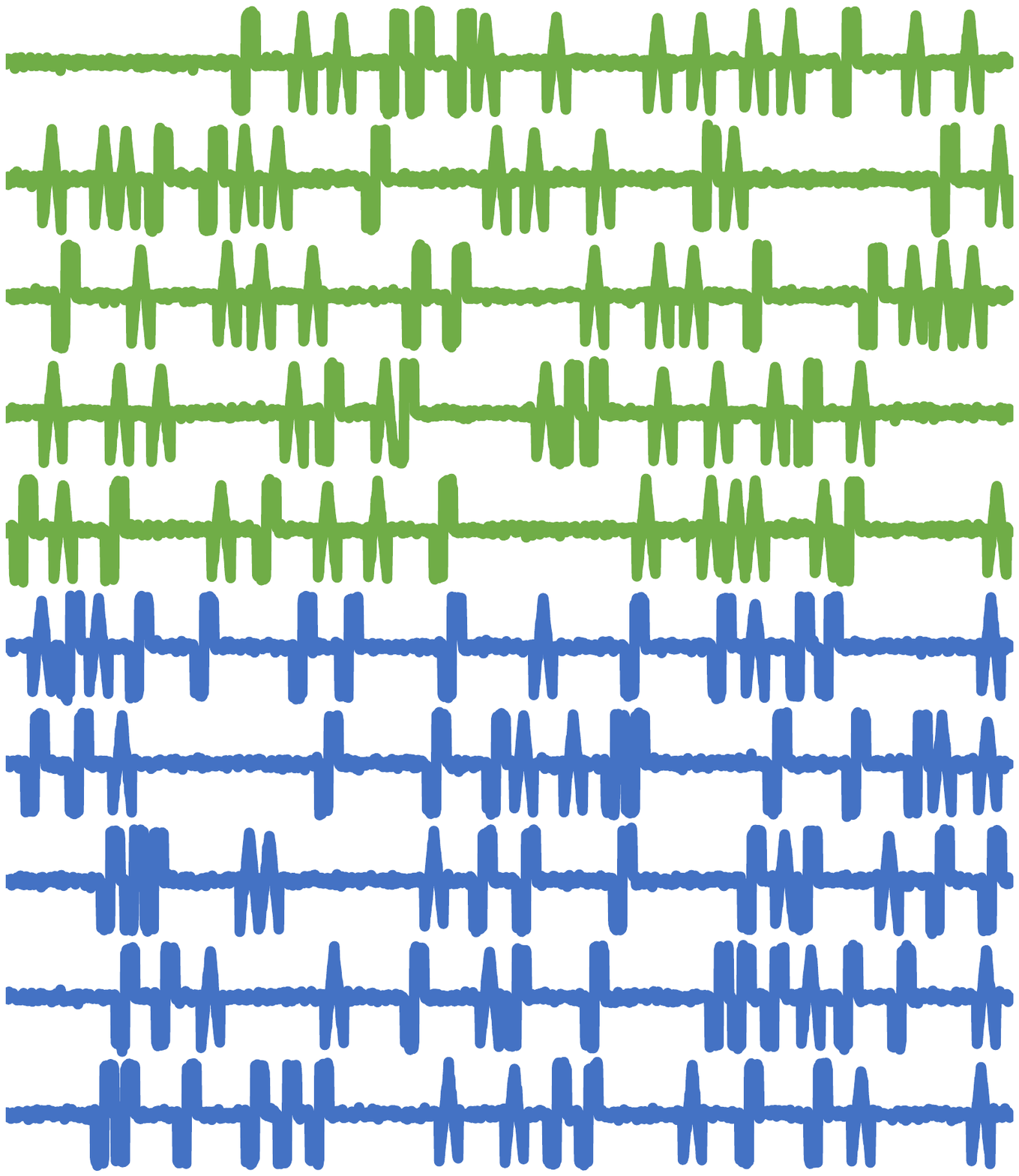}              	
&
       \includegraphics[height=6cm,trim={2cm 2cm 2cm 2cm},clip]{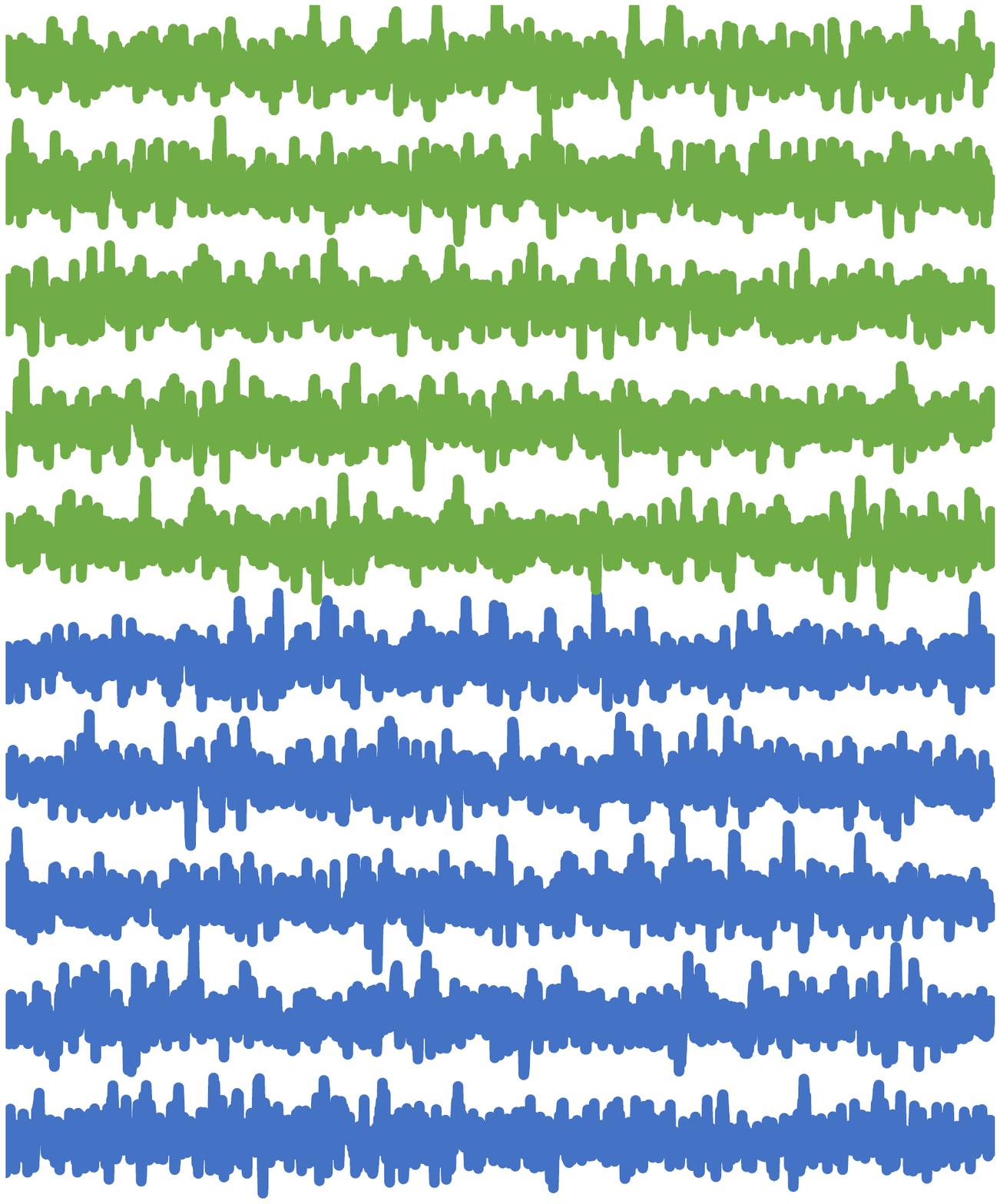}              	\\
       (a) & (b) \\
\end{tabular}
       \caption{Two examples of simulated dictionary data.  with a mixture of truncated head and shoulders and sine shapelets.  Figure (a) has low noise and Figure (b) has standard white noise. }
       \label{dictionaryExample}
\end{figure}

The basic assumption is that for some problems classes are defined by the frequency of occurrence of a shape rather than its presence or absence. To model this scenario, we include two shapes per series, and define class membership on the number of occurrences of these shapes. For example, one class may have ten triangles and five steps placed at random locations for each series of that class and the other have five triangles and ten step. Figure~\ref{dictionaryExample} gives some example series with low and standard noise.

\subsection{Prior Belief}
We have the following prior beliefs concerning performance on this data:
\begin{enumerate}
\item The random locations of the shapes should make the vector based classifiers little better than random;
\item Elastic classifiers may exploit the fact that series of one class are more likely to appear next to each other to perform quite well;
\item Interval classifiers may capture the repetition over patterns, so we think they will do better than elastic classifiers;
\item This data should confound shapelet approaches, because both types of shape are in both classes;
\item Depending on the actual shape, spectral classifiers may handle this data relatively well, because repeated shapes should be detected in the power spectrum.
\end{enumerate}

\subsection{Results}

\begin{figure}[!ht]
	\centering
       \includegraphics[width=12cm,trim={4cm 12cm 4cm 11.5cm},clip]{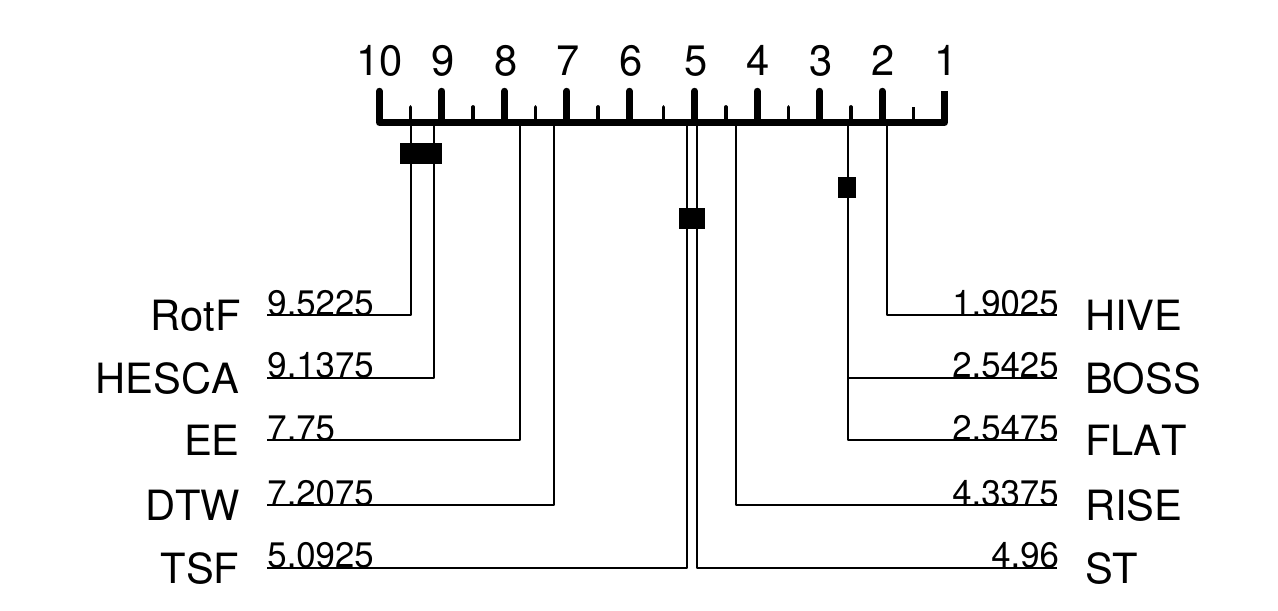}              	
       \caption{Critical difference diagram for ten classifiers over 200 dictionary simulations. }
       \label{dictCD}
\end{figure}

\begin{figure}[!ht]
	\centering
       \includegraphics[width=12cm,trim={3.5cm 10cm 3.5cm 10cm},clip]{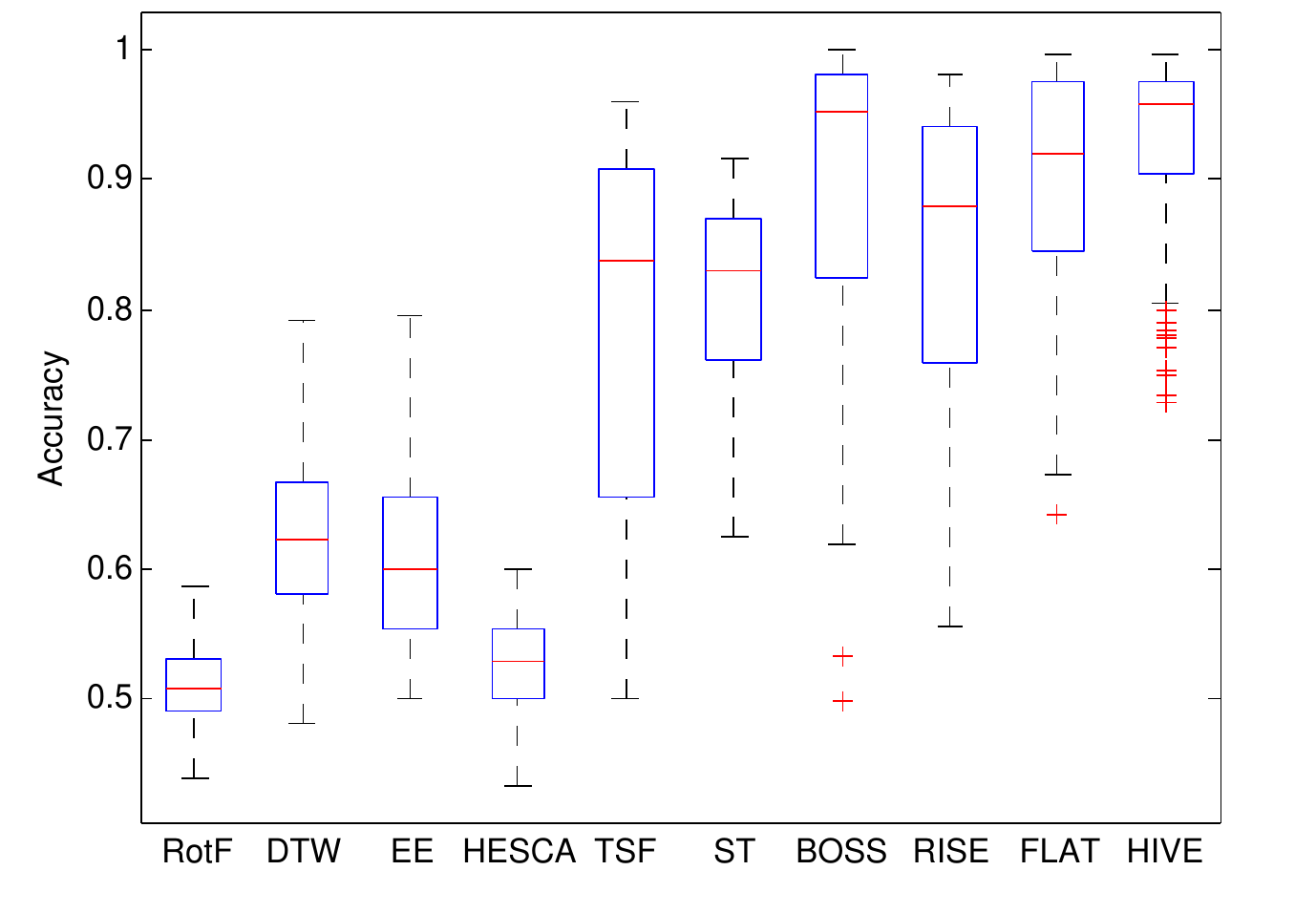}              	
       \caption{Box plots for ten classifiers over 200 dictionary simulations. }
       \label{dictBoxPlot}
\end{figure}

The parameters for the dictionary simulator are as follows
\begin{verbatim}
    int[] shapeletsPerClass={5,10};//Also defines the num classes by length
    int shapeLength=29;
    seriesLength =1500;
    int[] nosCases={200,200};
    double trainProp=0.1;
\end{verbatim}
The series are much longer in this simulation because we have many more patterns. The results are presented in Table~\ref{dictAcc} and Figures~\ref{dictCD} and ~\ref{dictBoxPlot}.These results can be recreated with the method \texttt{SimulationExperiment.runShapeletDictionaryExperiment()}.

\begin{table}
\centering
\caption{Mean Accuracy and rank of ten classifiers averaged over 200 different random Dictionary datsets.}
\label{dictAcc}
\begin{tabular}{cccc} \hline
Algorithm   & Mean Accuracy & Standard Error & Mean Rank\\ \hline
HIVECOTE    & 93.05\% & 0.44\%  & 1.90\\
BOSS        & 90.35\% & 0.72\%  & 2.54\\
FLATCOTE    & 90.21\% & 0.56\%  & 2.55\\
RISE        & 84.08\% & 0.84\%  & 4.34\\
ST-HESCA          & 81.37\% & 0.46\%  & 4.96\\
TSF         & 78.99\% & 0.99\%  & 5.09\\
DTW         & 62.78\% & 0.43\%  & 7.21\\
EE          & 60.65\% & 0.47\%  & 7.75\\
HESCA       & 52.54\% & 0.25\%  & 9.14\\
RotF        & 50.98\% & 0.20\%  & 9.52\\ \hline
\end{tabular}
\end{table}

\subsection{Post Experiment Observations}

Broadly, the results are in line with our expectations. \\

\noindent{\bf Vector based}.
Both RotF and HESCA are little better than random guessing.\\

\noindent{\bf Elastic}. Both DTW and EE are better than the vector classifiers and DTW is actually significantly better on these simulations than EE. We have no explanation for this, but it is not particularly relevant, as all the other classifiers are significantly better than both.\\

\noindent{\bf Shapelet and interval}. There is no difference between ST-HESCA and TSF in ranks, and both are recovering some of the discriminatory features. TSF has a much wider range of accuracies, which implies it does well when the intervals happen to overlap. The ST-HESCA result confounded our expectations. We believe the ST-HESCA does better than random because it finds shapelets that overlap more than one shape: a shapelet that contains two sine waves is more likely to be from a class with a predominance of sine waves.    \\

\noindent{\bf Spectral}. As expected, RISE does better than both TSF and ST-HESCA.\\

\noindent{\bf Dictionary}. BOSS is significantly better than the other single representation classifiers. \\

\noindent{\bf Combined}. Somewhat surprisingly, Flat-COTE is no different to BOSS and HIVE-COTE is significantly better than BOSS. This implies that the components are discovering discriminatory information that BOSS does not for some problems.

\section{ARMA Simulator \\ \texttt{ArmaModel} and \texttt{generateArmaData}}

\subsection{Prior Beliefs}
\begin{enumerate}
\item The nature of this data would make us believe that Vector based, Elastic, Interval, Shapelet and Dictionary  would be equally poor.
\item RISE would, we hope, be significantly more accurate than all other approaches. \\
\item If we are correct, the COTE classifiers will be combining one good classifier (RISE), with several poor ones. We would hope that it could detect this through cross validation, but suspect both Flat-COTE an HIVE-COTE may be significantly worse than RISE.
\end{enumerate}

\subsection{Results}
\begin{table}
\centering
\caption{Mean accuracy and rank of ten classifiers averaged over 200 different random ARMA datsets.}
\label{elasticAcc}
\begin{tabular}{l|c|c|c} \hline
Algorithm & Mean Accuracy & Standard Error & Mean Rank\\ \hline
RISE        & 81.82\% & 0.75\%  & 1.42\\
ST-HESCA          & 77.11\% & 0.91\%  & 2.98\\
HIVECOTE    & 77.09\% & 0.86\%  & 2.91\\
FLATCOTE    & 76.86\% & 0.83\%  & 2.99\\
EE          & 63.50\% & 0.79\%  & 5.78\\
TSF         & 60.80\% & 0.80\%  & 6.50\\
DTW         & 60.34\% & 0.70\%  & 6.80\\
BOSS        & 55.95\% & 0.64\%  & 8.04\\
HESCA       & 52.37\% & 0.44\%  & 8.80\\
RotF        & 52.14\% & 0.39\%  & 8.79\\ \hline
\end{tabular}
\end{table}

\begin{figure}[!ht]
	\centering
       \includegraphics[width=12cm,trim={4cm 12cm 4cm 11.5cm},clip]{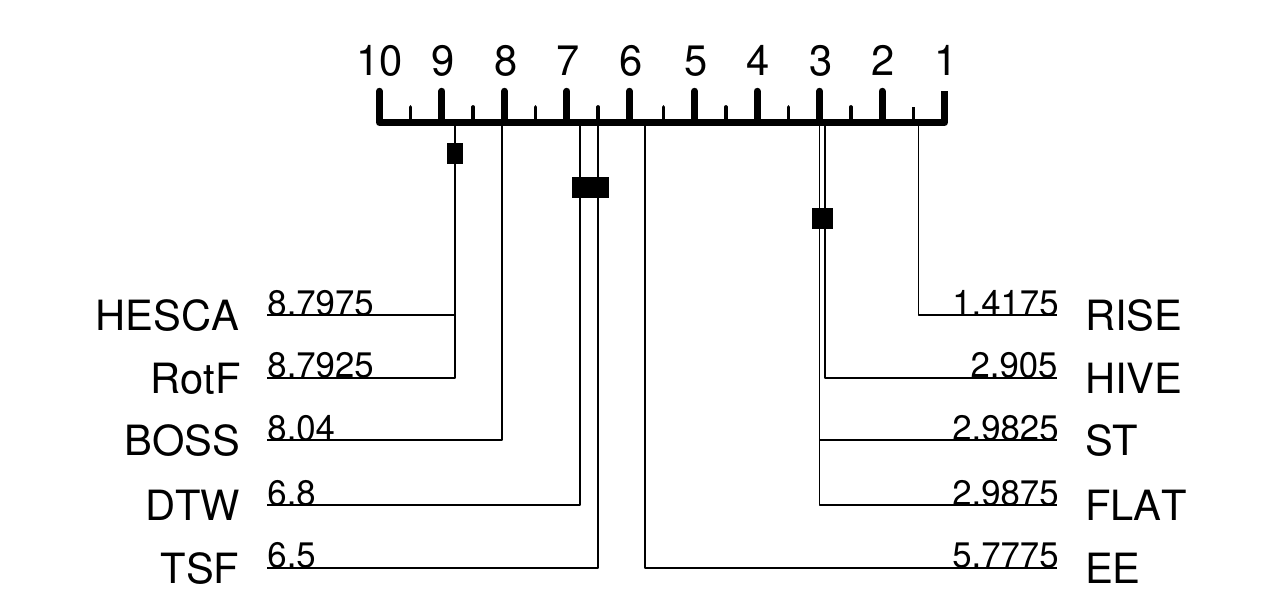}              	
       \caption{Critical difference diagram for ten classifiers over 200 arma simulations.}
       \label{armaCD}
\end{figure}

\begin{figure}[!ht]
	\centering
       \includegraphics[height=10cm,trim={2cm 8cm 2cm 8cm},clip]{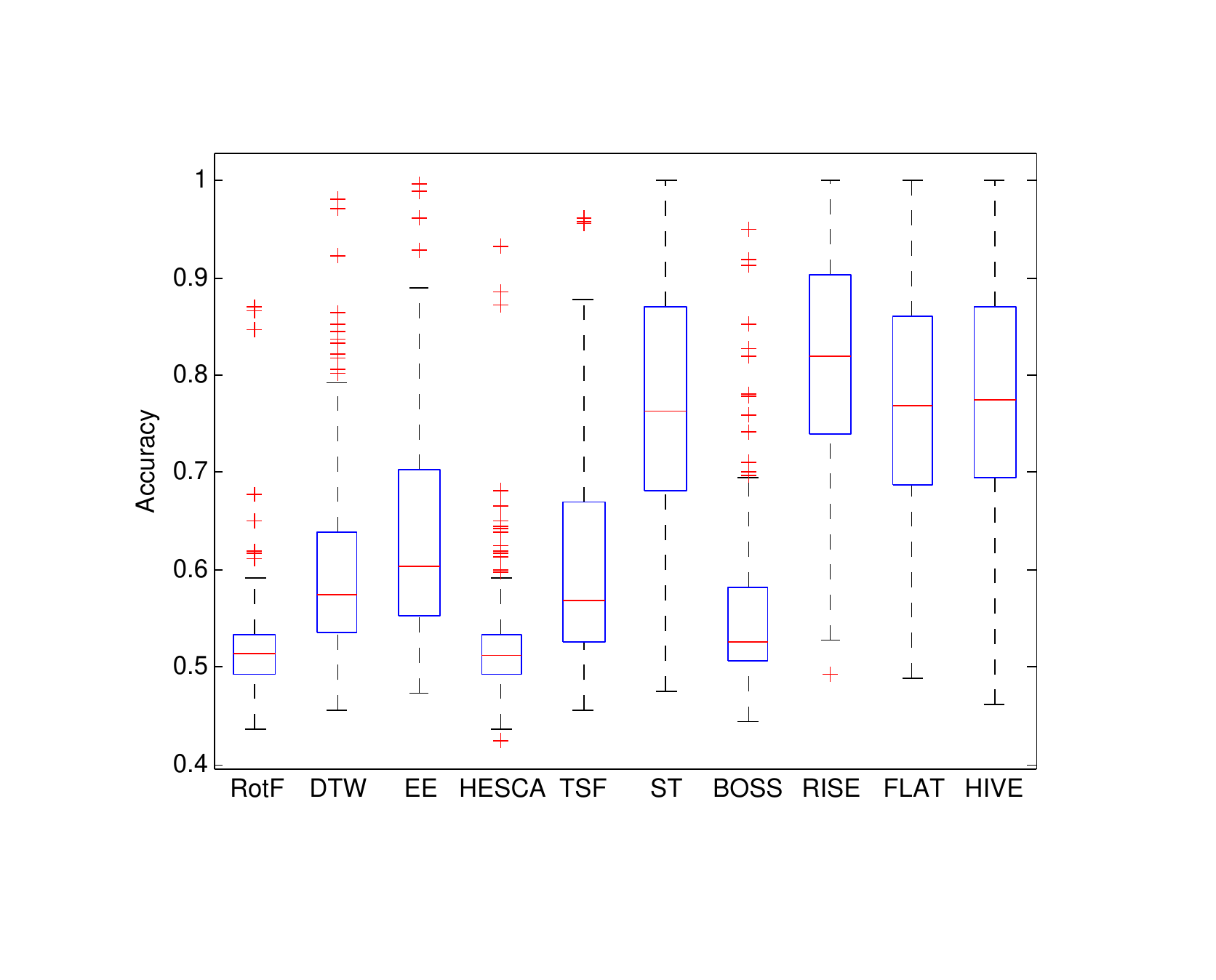}              	
       \caption{Box plots for ten classifiers over 200 ARMA simulations. }
       \label{armaBoxPlot}
\end{figure}

\subsection{Post Experiment Observations}

Broadly, the results are in line with our expectations. There are some anomalies though.\\

\noindent{\bf Vector based, dictionary, interval and elastic}. RotF and HESCA are equally poor, only just better than random guessing. BOSS is significantly better than the vector classifiers, but significantly worse than DTW and TSF, which form a clique. EE manages to do a little better, probably because of the presence of difference based classifiers, but is still not much better than random guessing.\\

\noindent{\bf Shapelet}.Mysteriously, ST-HESCA does significantly better than other approaches. ARMA data are characterised but short order autocorellations, so it is conceivable that ST-HESCA finds short shapelets that capture this.\\

\noindent{\bf Spectral}. As expected, RISE is significantly more accurate than all other approaches.\\

\noindent{\bf Combined}. Both Flat-COTE and HIVE-COTE are significantly worse than RISE. We believe
this results indicate two things. Firstly, ST-HESCA is not discovering discriminatory features that RISE is missing. It is simply providing a poor approximation of the RISE classifier. Secondly, we know that ST-HESCA is biased in its estimate of train cross validation accuracy, and we believe it is this that is reducing the quality of the COTE classifiers.

\section{All Combined}
\label{all}
Suppose we do not know what simulator the data originated from. What would be the best approach then? We model this scenario by pooling all our results into one experiment with 1000 resamples.
\subsection{Prior Beliefs}
\begin{enumerate}
\item We expect HIVE-COTE to be significantly better than Flat-COTE and Flat-COTE to be significantly better than all other approaches
\item We expect ST-HESCA to be the best of the components, followed by BOSS. 
\end{enumerate}

\subsection{Results}
\begin{table}
\centering
\caption{Mean accuracy and rank of ten classifiers averaged over 1000 different random datsets.}
\label{elasticAcc}
\begin{tabular}{l|c|c|c} \hline
Algorithm & Mean Accuracy & Standard Error & Mean Rank\\ \hline
RISE        & 81.82\% & 0.75\%  & 1.42\\
ST-HESCA          & 77.11\% & 0.91\%  & 2.98\\
HIVECOTE    & 77.09\% & 0.86\%  & 2.91\\
FLATCOTE    & 76.86\% & 0.83\%  & 2.99\\
EE          & 63.50\% & 0.79\%  & 5.78\\
TSF         & 60.80\% & 0.80\%  & 6.50\\
DTW         & 60.34\% & 0.70\%  & 6.80\\
BOSS        & 55.95\% & 0.64\%  & 8.04\\
HESCA       & 52.37\% & 0.44\%  & 8.80\\
RotF        & 52.14\% & 0.39\%  & 8.79\\ \hline
\end{tabular}
\end{table}

\begin{figure}[!ht]
	\centering
       \includegraphics[width=12cm,trim={4cm 12cm 4cm 11.5cm},clip]{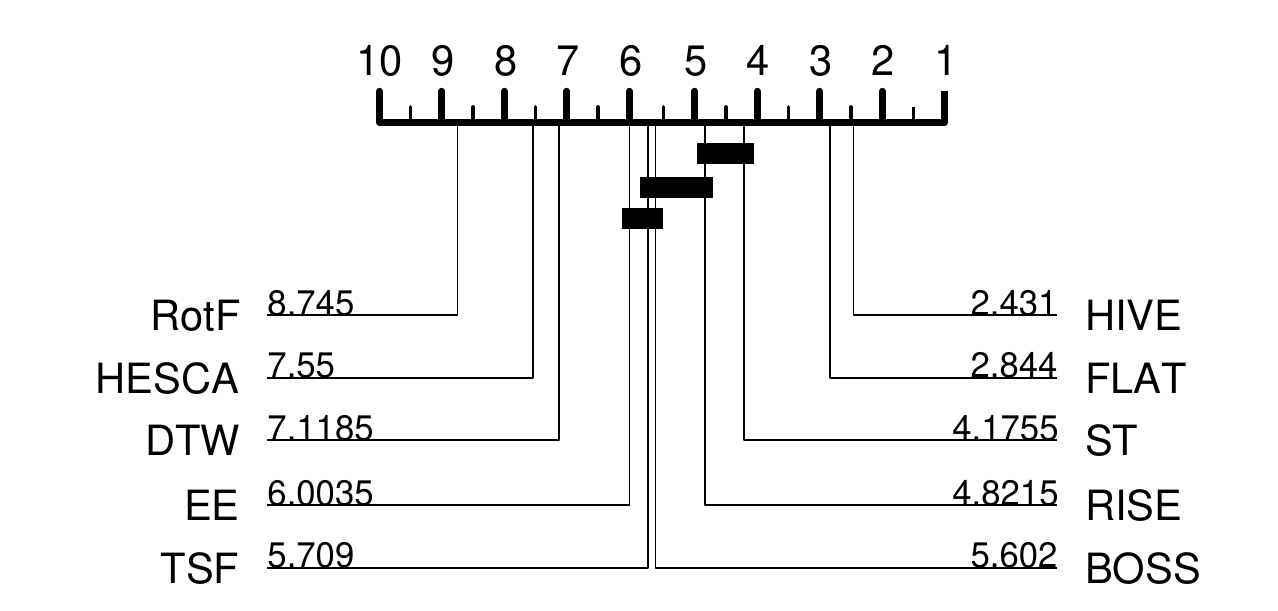}              	
       \caption{Critical difference diagram for ten classifiers over 1000 simulations. CHECK CLIQUES}
       \label{allCD}
\end{figure}

\begin{figure}[!ht]
	\centering
       \includegraphics[height=10cm,trim={2cm 8cm 2cm 8cm},clip]{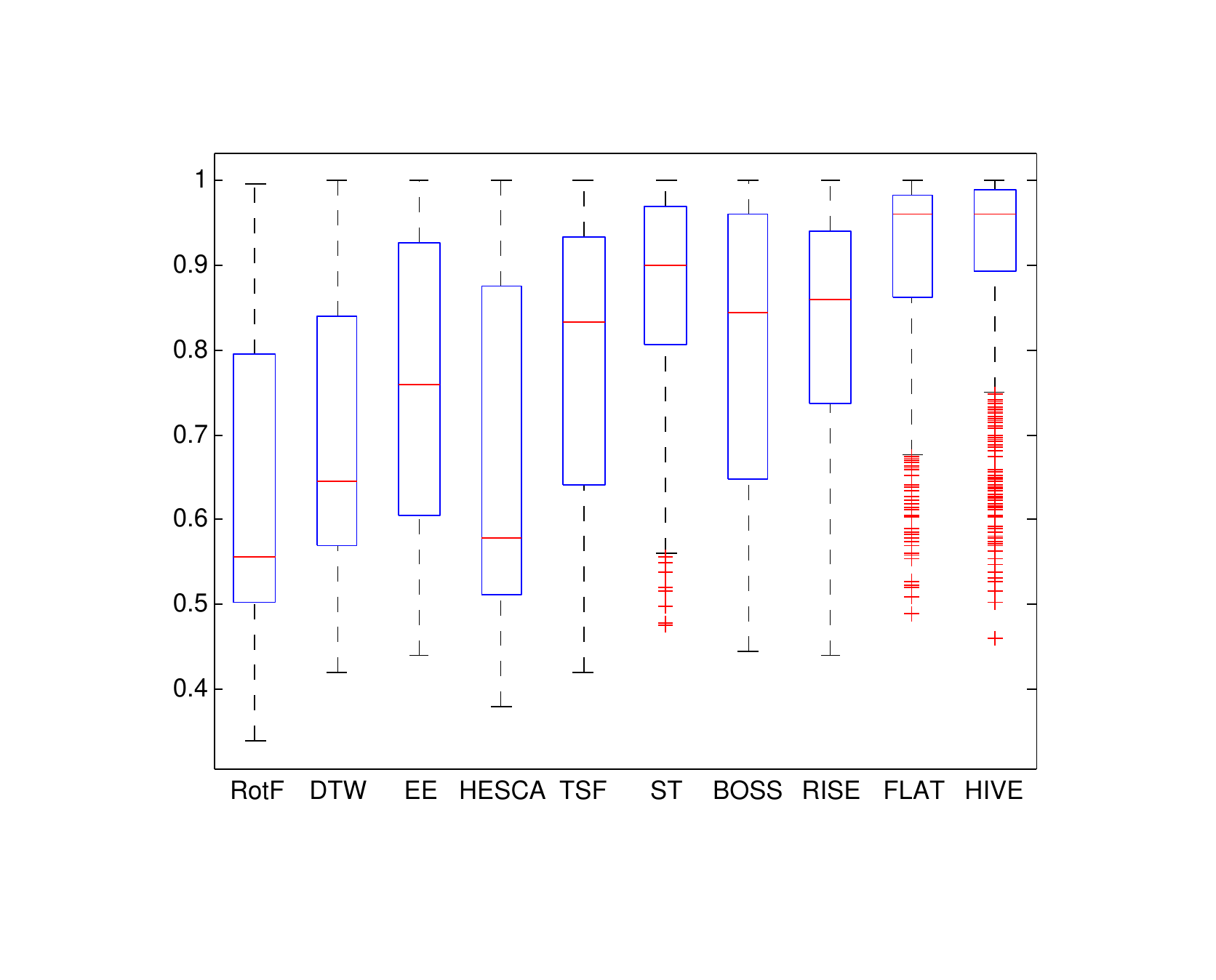}              	
       \caption{Box plots for ten classifiers over 1000 simulations. }
       \label{armaBoxPlot}
\end{figure}
\subsection{Post Experiment Observations}
Broadly, the results are in line with our expectations. \\

\noindent{\bf Vector based, Elastic, Interval, Shapelet and Dictionary}. The nature of this data would make us believe that all of these approaches would be equally poor. \\

\noindent{\bf Spectral}. RISE would, we hope, be significantly more accurate than all other approaches. \\

\noindent{\bf Combined}. If we are correct, the COTE classifiers will be combining one good classifier (RISE), with several poor ones. We would hope that it could detect this through cross validation, but suspect both Flat-COTE an HIVE-COTE may be significantly worse than RISE.

\section{Conclusions}

This sequence of simulation experiments was primarily conducted to test the hypothesis that HIVE-COTE produces significantly more accurate classifiers. Overall, we think these results support our core hypothesis that HIVE-COTE is the best currently available technique for TSC. It is significantly more accurate than all competitors on the UCR/UEA repository problems (see~\cite{lines16hive}, significantly more accurate than all classifiers when the data is randomly selected from one of five simulators and significantly better than, or not significantly worse than, the best other approach on three out of five of the individual simulators. The two it is significantly worse than the best approach are Interval (HESCA wins, HIVE second) and Spectral (RISE wins, HIVE second). HIVE-COTE represents the state-of-the-art in TSC, and our efforts are now directed towards making it faster. Our secondary goal is to begin the process of better understanding why one classifier is better on certain data types than others, and these experiments have highlighted several issues we think worthy of further investigation.

\begin{enumerate}
\item Why is HESCA significantly better than RotF on the simulators, and can we find a general case in classification where the HESCA approach is better than standard classifiers?
\item Why is EE better than DTW on the interval data, and can we better understand how the components of EE interact?
\item Does TSF actually offer any benefit over EE, or is it simply duplicating its performance?
\item If TSF and interval approaches are distinct, what type of data are they best for and is there scope for a better interval approach?
\item What data characteristics are optimal for dictionary based approaches and why does BOSS do so much better than the alternatives?
\end{enumerate}

\end{document}